\documentclass[10pt,twocolumn,letterpaper]{article}

\usepackage[accsupp]{axessibility}  

\usepackage{cvpr}              

\usepackage{graphicx}
\usepackage{amsmath}
\usepackage{amssymb}
\usepackage{booktabs}

\usepackage{booktabs}
\usepackage{multirow}
\usepackage{nicefrac}      
\usepackage{algorithm}
\usepackage{algpseudocode}
\usepackage{caption}
\usepackage{colortbl}

\usepackage{amsmath,amssymb,amsfonts}

\newcommand{\vct}[1]{\ensuremath{\mathbf{#1}}}

\newcommand{\argmin}{\operatornamewithlimits{\arg\,\min}}

\usepackage{textcomp}
\usepackage{gensymb}

\usepackage{bm}

\definecolor{Gray}{gray}{0.92}

\usepackage[pagebackref,breaklinks,colorlinks]{hyperref}

\usepackage[capitalize]{cleveref}
\crefname{section}{Sec.}{Secs.}
\Crefname{section}{Section}{Sections}
\Crefname{table}{Table}{Tables}
\crefname{table}{Tab.}{Tabs.}
\renewcommand{\ie}{{i.e.,}\xspace}
\renewcommand{\eg}{{e.g.,}\xspace}

\def\cvprPaperID{11039} 
\def\confName{CVPR}
\def\confYear{2023}

\begin{document}

\title{Introducing Competition to Boost the Transferability of \\Targeted Adversarial Examples through Clean Feature Mixup}

\author{Junyoung Byun \quad Myung-Joon Kwon \quad Seungju Cho \quad Yoonji Kim\quad Changick Kim \\
Korea Advanced Institute of Science and Technology (KAIST)\\
{\tt\small \{bjyoung, kwon19, joyga, yoonjikim, changick\}@kaist.ac.kr}
}
\maketitle

\begin{abstract}
Deep neural networks are widely known to be susceptible to adversarial examples, which can cause incorrect predictions through subtle input modifications. These adversarial examples tend to be transferable between models, but targeted attacks still have lower attack success rates due to significant variations in decision boundaries. To enhance the transferability of targeted adversarial examples, we propose introducing competition into the optimization process. Our idea is to craft adversarial perturbations in the presence of two new types of competitor noises: adversarial perturbations towards different target classes and friendly perturbations towards the correct class. With these competitors, even if an adversarial example deceives a network to extract specific features leading to the target class, this disturbance can be suppressed by other competitors. Therefore, within this competition, adversarial examples should take different attack strategies by leveraging more diverse features to overwhelm their interference, leading to improving their transferability to different models. Considering the computational complexity, we efficiently simulate various interference from these two types of competitors in feature space by randomly mixing up stored clean features in the model inference and named this method Clean Feature Mixup (CFM). Our extensive experimental results on the ImageNet-Compatible and CIFAR-10 datasets show that the proposed method outperforms the existing baselines with a clear margin. Our code is available at
\url{https://github.com/dreamflake/CFM}.\end{abstract}

\begin{figure*}[t]
         \centering
         \includegraphics[width=0.8\textwidth,trim={0cm 0cm 0cm 0cm},clip]{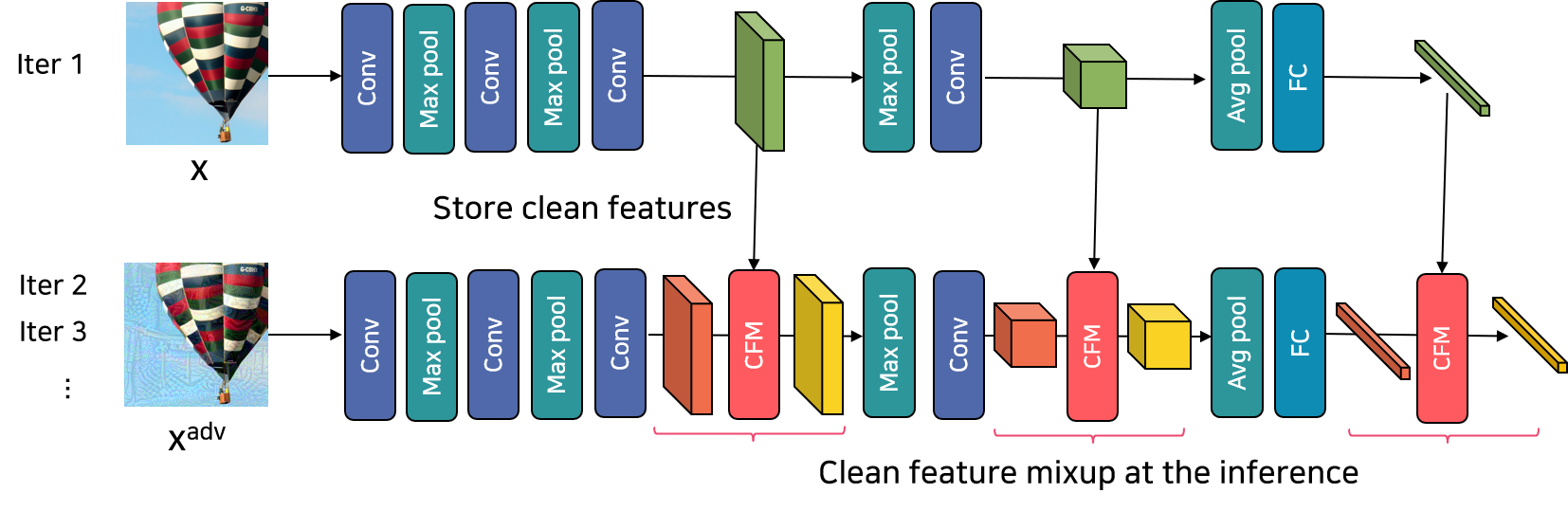}
         \vspace{-0.2cm}
         \caption{Overview of the Clean Feature Mixup (CFM) method.}
        \label{fig:fig1}
        \vspace{-0.2cm}
\end{figure*}

\section{Introduction}
\label{sec:intro}
Although deep neural networks have excelled in various computer vision tasks such as image classification \cite{he2016deep,huang2017densely} and object detection \cite{redmon2016you,liu2016ssd}, they are vulnerable to maliciously crafted inputs called \textit{adversarial examples} \cite{goodfellow2014explaining,zhao2021success}. These adversarial examples are generated by optimizing imperceptible perturbations to mislead a model to incorrect predictions. Intriguingly, these adversarial examples tend to be transferable between models, and this unique characteristic allows adversaries to attempt adversarial attacks on a black-box model without knowing its interior. However, targeted adversarial attacks, which have a specific target class, still have lower attack success rates due to significant differences in decision boundaries \cite{li2020towards,zhao2021success}. Nevertheless, targeted attacks can pose more serious risks as they can deceive models into predicting a specific harmful target class. Therefore, preemptive research on developing a novel transfer-based attack is crucial because it can assist service providers in preparing their models for these forthcoming risks and evaluating their models' robustness. 

In this work, we aim to further improve the transferability of targeted adversarial examples by introducing competition into their optimization. Our approach involves crafting adversarial perturbations in the presence of two new types of noises: \textit{(a) adversarial perturbations towards different target classes}; and \textit{(b) friendly perturbations towards the correct class}. With these competitors and a source model, even if an adversarial example deceives the source model into extracting certain features that lead to the target class, this disturbance may be suppressed by interference from competitors. Consequently, adversarial perturbations should take various attack strategies, leveraging a wider range of features to overcome interference, which enhances their transferability to different models. In the following, we will further discuss why employing a diverse set of features for attack can boost the transferability of targeted adversarial examples.

In image classification, deep learning models extract a variety of features from images across multiple layers and comprehensively evaluate them to calculate prediction probabilities for each class. As numerous features can contribute to the final output, even when two images are recognized as the same class, the contributing features can significantly differ. Taking this into account, optimizing adversarial examples to utilize as many distinct feature combinations as possible would effectively enhance their transferability. 

Conversely, in existing frameworks, an adversarial example may be optimized to intensely distract a limited number of features identified in the early stages. However, the target model, unlike the source model, might be insensitive to such feature distractions, leading to the failure of transfer-based attacks.

Meanwhile, if we model the competitors as noises that should also be optimized, they require additional backward passes, substantially increasing the computational burden. To address this challenge and enhance interference diversity, we propose \textit{Clean Feature Mixup} (CFM), a method that efficiently mimics the behaviors of the two types of perturbations in feature space by mixing stored clean features of the images within a batch. A detailed description of their similarities can be seen in Section \ref{sec:cfm_principle}.

The overview of the proposed CFM method is illustrated in Fig. \ref{fig:fig1}. Specifically, this method converts a pre-trained source model by attaching our specially designed CFM modules to convolution and fully-connected layers. After that, the attached CFM modules randomly mix the features of the clean images (\ie \textit{clean features}) with current input features at each inference. This process can effectively mitigate the overfitting of adversarial examples in their optimization by preventing them from focusing on particular features in their targeted attacks on the source model. Unlike many existing techniques \cite{lin2019nesterov,wang2021enhancing,wang2021admix} that significantly increase the computational cost by multiplying the required number of forward/backward passes, CFM adds just one additional forward pass for storing clean features and requires a marginal amount of computation for feature mixup at each inference. 

Our contributions can be summarized as follows:
\setlength\itemsep{-0.1em}
\begin{itemize} 

 \item We propose the idea of introducing competition into the optimization of targeted adversarial examples with two types of competitor noises to encourage the utilization of various features in their attacks, ultimately boosting their transferability.
 
    \item Motivated by the above idea, we propose the Clean Feature Mixup (CFM) method to improve the transferability of adversarial examples. This method efficiently simulates the competitor noises by randomly mixing up stored clean features of the images in a batch.

    \item We performed extensive experiments with 20 models, including four defensive models and five Transformer-based classifiers. Our experimental results on the ImageNet-Compatible and CIFAR-10 datasets demonstrate that CFM outperforms state-of-the-art baselines.
\end{itemize}

\section{Background}
In this section, we briefly overview the background of adversarial attacks, various techniques to improve adversarial transferability, and defensive models against them.

\subsection{Adversarial Attacks}
\label{sec:adv_atack}
Given a clean image $\vct x$ with its true label $y$ and a source model $f$, adversarial attacks inject inconspicuous noise $\bm{\delta}$ to craft adversarial examples $\vct x^{adv} = \vct x + \bm{\delta}$, causing the model to make erroneous predictions. The imperceptibility constraint of adversarial perturbation can be set differently, but the $\ell_\infty$ norm is widely used in existing works \cite{xie2019improving,zou2020improving,dong2019evading,wang2021enhancing,zhao2021success, byun2022improving}.
Under the white-box setting, adversaries can utilize the gradient of the loss function (\eg the cross-entropy loss) with respect to the input image for the optimization of an $\ell_\infty$-norm constrained targeted adversarial example towards a given target class $y_{t}$ as follows:
\begin{align}%
\label{eqn:adv_eqn}
    \argmin_{\vct x^{adv}} \mathcal{L}(f(\vct x^{adv}),y_{t}) \; s.t \; ||\vct x-\vct x^{adv}||_{\infty} \leq \epsilon,
\end{align}
where $\mathcal{L}$ is the adversary's objective loss for targeted attacks. One of the most fundamental attack methods for the above optimization is the Fast Gradient Sign Method (FGSM) \cite{goodfellow2014explaining}, which utilizes the sign of the gradient. For targeted attacks, it updates $\vct x$ towards the direction of minimizing the classification loss to the given target class $y_{t}$ in a single step as follows:
\begin{equation}
    \vct x^{adv} = \vct x - \epsilon \cdot \text{sign}(\nabla_{\vct x} \mathcal{L}(f(\vct x),y_{t})),
    \label{eqn:adv}
\end{equation}
where $\epsilon$ represents the perturbation bound. It can be further optimized by updating the image iteratively with a smaller step size, $\eta$, as in Iterative-FGSM \cite{kurakin2016adversarial}. 
\subsection{Transfer-based Black-Box Attacks}
Under the black-box setting, the target model's interior cannot be accessed, so the gradient of the image cannot be directly computed via the back-propagation technique. Therefore, adversaries need to craft adversarial examples on white-box surrogate source models that mimic the target model's function. After that, the attackers can attempt black-box attacks by feeding the generated adversarial examples to the target model.

However, the transfer success rate varies significantly depending on the difference between the source and target models, such as architectural differences. Therefore, for successful targeted attacks on black-box models, it is essential to improve the transferability of adversarial examples generated on surrogate models by preventing them from overfitting the source models. To this end, various techniques have been proposed to improve transferability based on the fundamental adversarial attacks explained in Section \ref{sec:adv_atack}. These techniques include input diversification  \cite{xie2019improving,zou2020improving,byun2022improving}, gradient stabilization \cite{dong2018boosting,lin2019nesterov}, and use of different loss functions \cite{li2020towards,zhao2021success,wang2021feature,huang2018intermediate}. In the following, we briefly introduce these approaches. 

One of the representative methods for input diversification is the Diverse-Inputs (DI) method \cite{xie2019improving}. For each inference in iterative optimization, it randomly expands and pads the image with the probability $p$. The Resized-Diverse-Inputs (RDI) method \cite{zou2020improving} extends the DI technique by shrinking the expanded image to its original size at the end of the DI transform. Unlike DI, RDI always applies the image transform (\ie $p=1$). 

The Translation-Invariant (TI) attack \cite{dong2019evading} blurs image gradients, approximating a weighted average of gradients from a set of translated images within a certain range. This technique provides a degree of translation invariance to the adversarial examples, making them more transferable between models.

The Admix \cite{wang2021admix} further improves the transferability by mixing different images in the image domain. Specifically, according to the official implementation, Admix takes randomly shuffled images of the current batch, diminishing their pixel values by multiplying a mixing weight $w$, and adds them to the batch's images. 
Admix repeats the above addition-based mixup for $N$ times and computes the average gradients.

The Object-based Diverse Input (ODI) method \cite{byun2022improving} is a recent technique that naturally diversifies inputs. This approach draws an input image on the surface of a randomly chosen 3D object and renders this painted object in various rendering environments. Empirical results demonstrate that ODI significantly improves the transferability of targeted adversarial examples, achieving state-of-the-art performance.

Stabilizing image gradients is another approach that can improve adversarial transferability by preventing adversarial examples from falling into local optima. The Momentum Iterative FGSM (MI-FGSM) \cite{dong2018boosting} incorporates a momentum term in the iterative attacks. The Variance Tuning (VT) method \cite{wang2021enhancing} highlights gradient variance, the difference between the image's gradients and the average gradients of adjacent images. By minimizing gradient variance, VT can stabilize the update direction in the optimization process. Similarly, the Scale-Invariant (SI) attack method \cite{lin2019nesterov} scales down the pixel values of the input image in several steps and computes the gradients from the set of images. These techniques help to alleviate the overfitting of adversarial examples and thus improve transferability.

As another direction to improve transferability, several studies \cite{li2020towards,zhao2021success} have suggested different loss functions for targeted attacks. Zhao \etal \cite{zhao2021success} point out that previous works use insufficient iterations to reach the optimal point in crafting adversarial examples. They empirically show that with large enough steps, significant performance improvement can be achieved with the following simple logit loss to increase the target class's logit.
\begin{equation}
    \mathcal{L}_{logit}(f(\vct x^{adv}),y_{t})=-\ell_{t}(f(\vct x^{adv})),
    \label{eqn:logit}
\end{equation}
where $\ell_{t}$ is the logit value corresponding to the target class $y_{t}$.

\begin{figure*}[t]
         \centering
         \includegraphics[width=0.8\textwidth,trim={0cm 0cm 0cm 0cm},clip]{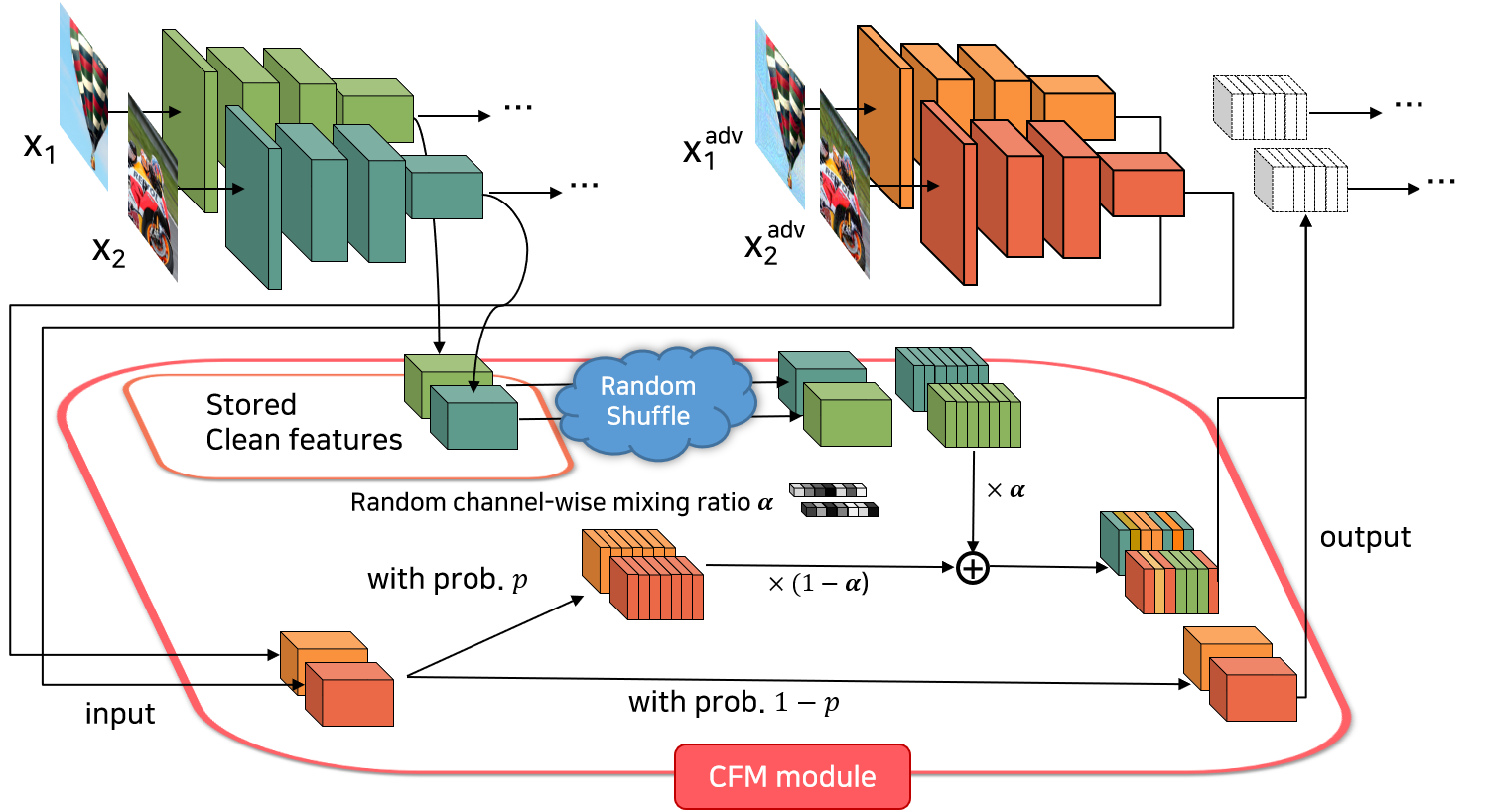}
         \caption{The detailed illustration of the internal process of the CFM module.}
        \label{fig:fig2}
        \vspace{-0.3cm}
\end{figure*}
\subsection{Defensive Models}
Several studies have also been conducted to build more robust models against transfer-based black-box attacks.
One of the representative methods for defending against adversarial attacks is adversarial training \cite{madry2017towards,tramer2018ensemble}, which directly utilizes adversarial examples in training models.
Another effective defense strategy is constructing an ensemble of individual networks, where adversaries should fool multiple models in the ensemble instead of a single model, thus improving robustness. To this end, an ensemble model is trained using the usual cross-entropy loss and some regularizing terms to consider the interaction among individual networks. Pang \etal \cite{pang2019improving} introduce the Adaptive Diversity Promoting (ADP) regularizer, which enhances the diversity among non-maximal predictions of individual members to make adversarial examples more difficult to transfer among them.
Kariyappa \etal 
\cite{kariyappa2019improving} propose the Gradient Alignment Loss (GAL) regularizer to misalign loss gradients, reducing the dimensionality of the shared adversarial subspace.
DVERGE \cite{yang2020dverge} trains sub-models to utilize a distinct set of features, isolating the adversarial vulnerability in each sub-model.

\section{Clean Feature Mixup}
The proposed Clean Feature Mixup is a method designed to enhance the transferability of targeted adversarial examples by efficiently emulating two types of competitors in the optimization process, as motivated in Section \ref{sec:intro}. In this section, we will first describe the implementation of CFM and then explain how CFM can simulate the two competitor noises in Section \ref{sec:cfm_principle}.

As an overview, the proposed CFM technique transforms the feature maps in their optimization to prevent adversarial examples from overfitting the source model. 
However, the feature space is much broader than the image space, and each model can have a different architecture, making it challenging to decide where and how to transform the features. Naively applying conventional image transforms to feature maps may excessively impede the optimization due to the significant domain gap between images and features. Instead, we transform the features by randomly mixing clean features with the input features. Specifically, for convolution and fully connected layers, it mixes the layers' outputs (\ie  features) with the stored clean features via linear interpolation \cite{zhang2018mixup}. 

To do this, it is necessary to store clean features in memory to mix them at inference. However, the structures of deep neural networks are not uniform, and they may have different modules. Our particular interest is to devise an off-the-shelf implementation to mix clean features while minimizing the effort of modifying the existing codes for the network architectures. Taking this into consideration, we design the CFM module that performs the above two functions (\ie storing clean features and mixing them in the input features) and simply append the CFM modules to deeper convolution and fully connected layers as shown in Fig. \ref{fig:fig1}. In the following, we describe each function of the CFM module in more detail.

\subsection{Storing Clean Features}
To mix the clean features in model inferences, the CFM modules require storing the clean features in the memory at first. To do this, it first converts a pre-trained source model $f$ to the CFM-attached model $f'$ by attaching the CFM modules to selected \textit{conv} layers and all \textit{fc} layers, and passes the clean image $\vct x$ to the converted model $f'$. Each CFM module stores the clean features in its memory at this first inference.

To avoid excessive disturbance in the optimization process, we do not append CFM modules to all convolution layers but only to deeper layers where the output size is significantly smaller than the input image. Mixing larger-sized, low-level features can cause excessive disturbance in the optimization process as they can vary significantly based on the input transforms.
Therefore, we apply CFM modules only when the output's spatial size is less than or equal to $\frac{1}{16}$ of the original input size, which typically occurs after passing two pooling layers. We store pre-activated features since features can lose some information after passing through ReLU activations and apply feature mixup for these features.

\subsection{Mixing Stored Clean Features with Input Features}
As an overview, the internal process of a CFM module at the inference is depicted in Fig. \ref{fig:fig2}. We will explain the internal functions in order. 

\noindent\textbf{Stochastic activation.} Deep neural networks usually have tens to hundreds of layers, and applying clean feature mixup in all these layers at once can disrupt the inference process excessively. To address this issue, each CFM module stochastically applies the clean feature mixup with a probability $p$. This allows features to be mixed in a certain percentage of the total layers while maintaining randomness. Furthermore, this approach helps to obtain consistent performance gains regardless of the number of layers, thus reducing the effort required for hyperparameter tuning. From the perspective of competitor noises, this stochastic approach causes the influence of competitors to occur at random layers of the network.

\noindent\textbf{Random feature shuffle.} 
We also randomly shuffle the stored clean features on an image-wise basis within a batch. This allows the clean features of the image itself or those of another image to be mixed. Consequently, this enables the selection of competitor noises as either adversarial perturbations towards another target class or friendly perturbations towards the correct class, which is described in detail in Section \ref{sec:cfm_principle}.

\noindent\textbf{Random channel-wise mixing ratio.}
Each CFM module mixes the stored clean features and the input features via linear interpolation, and for more randomness, the mixing ratio is randomly sampled for each channel. This allows the effects of competitor noises to vary arbitrarily across channels, enabling more diverse interference.

Mathematically, given a batch of $B$ images, each CFM module stores $B$ clean feature maps, denoted as $\vct f^c_1, \dots, \vct f^c_B$, where $\vct f^c_i \in \mathbb{R}^{C\times H \times W}$ for $i = 1, \dots, B$. The variables $C$, $H$, and $W$ represent the number of channels, height, and width of the feature map, respectively. Then, each CFM module randomly mixes them with the input feature maps $\vct f_1, \dots, \vct f_B \in \mathbb{R}^{C\times H \times W}$ at each inference as follows:
\begin{equation}
    \vct f'_i=(1-\bm{\alpha}_i) \odot \vct f_i  + \bm{\alpha}_i \odot \vct f^c_{s_i},~i=1, \dots, B,
\end{equation}
where $\odot$ denotes element-wise multiplication and $s_i$ is the $i$-th element of randomly shuffled indices (for image-level feature shuffling) and $\bm{\alpha}_i\in \mathbb{R}^{C\times 1 \times 1}$ is the random channel-wise mixing ratio vector for the $i$-th image, where $\bm{\alpha}_i \sim \mathcal{U}(0,\alpha_{max})$, and $0\leq \alpha_{max}\leq 1$. The channel-wise mixing ratios $\{\bm{ \alpha}_1, \bm{ \alpha}_2, \dots, \bm{ \alpha}_B\}$ are sampled at each inference.

\subsection{How Can CFM Improve the Transferability of Targeted Adversarial Examples?}
\label{sec:cfm_principle}
The CFM modules randomly mix clean features of the image itself or another image into the input features, and this interference can further improve the transferability of the targeted adversarial examples in the following ways.

First, when an image's own clean features are mixed, this mixup suppresses the feature disturbance caused by the current targeted adversarial perturbations and guides the model's prediction back to the true class. In other words, it has the \textit{opposite effect} on the targeted adversarial attack, encouraging adversarial perturbations to explore alternative feature disturbances for successful attacks.

Second, when clean features of another image are mixed, this introduces the effect of \textit{targeted attacks on a different target class}\footnote{Exceptionally, if the shuffled features to be mixed come from an image of the same class, it has the opposite effect on the targeted attack.}. Consequently, the adversarial examples should be optimized to induce the model to predict the given target class in the presence of other targeted attacks on different classes, prompting the adversarial perturbations to explore robustly adversarial feature disturbances to succeed in the attacks.

In summary, the interference from clean feature mixup can \textit{effectively} and \textit{efficiently} mitigate overfitting in the optimization of adversarial examples by preventing them from concentrating on specific features during their targeted attacks on the source model. The CFM method is compatible with many existing attack methods, and as an example, the pseudo-codes of the CFM-RDI-MI-TI method are described in Appendix.

\section{Experiments}
\subsection{Experimental Settings}
\noindent\textbf{Datasets.} Following previous works~\cite{li2020towards,zhao2021success,byun2022improving}, we utilized the widely used ImageNet-Compatible dataset\footnote{\url{https://github.com/cleverhans-lab/cleverhans/tree/master/cleverhans_v3.1.0/examples/nips17_adversarial_competition/dataset}}, which was released for the NIPS 2017 adversarial attack challenge. It has 1,000 $299\times299$-sized images with their true and target classes for targeted attacks. 
We also leveraged the CIFAR-10 dataset \cite{krizhevsky2009learning} for targeted attacks on defensive models against transfer-based attacks. Specifically, we randomly sampled 1,000 images (100 images per class) from the test set and performed targeted attacks on randomly chosen incorrect target classes.

\noindent\textbf{General settings.}
Most of our experimental settings followed the recent study \cite{zhao2021success, byun2022improving}. Specifically, we employed the commonly used $\ell_\infty$-norm perturbation constraint with $\epsilon=16/255$ and set the step size $\eta=2/255$ for the iterative attacks following \cite{zhao2021success}. All the methods, including CFM and baselines, optimize the adversarial examples based on the simple logit loss \cite{zhao2021success, byun2022improving}. To give sufficient iterations to optimize adversarial examples, we set the total iterations $T$ to 300, which is also used in \cite{zhao2021success,byun2022improving}.

\begin{table*}[!t]
\centering
\resizebox{0.9\textwidth}{!}{%
\begin{tabular}{lccccccccccc}
\toprule[0.15em]
\textbf{\begin{tabular}[c]{@{}l@{}}Source : RN-50\end{tabular}} & \multicolumn{10}{c}{Target model} &  \\\cmidrule(l){2-11}
Attack & VGG-16 & RN-18 & RN-50 & DN-121 & Xcep & MB-v2 & EF-B0 & IR-v2 & Inc-v3 & Inc-v4 & Avg. \\ \midrule
DI & 62.5 & 56.6 & \textbf{98.9} & 72.3 & 5.7 & 28.2 & 29.3 & 4.5 & 9.2 & 9.9 & 37.7 \\
RDI & 65.4 & 71.8 & 98.0 & 81.3 & 13.1 & 46.6 & 46.6 & 16.8 & 30.7 & 23.9 & 49.4 \\
SI-RDI & 70.5 & 79.8 & 98.8 & 88.9 & 29.5 & 56.2 & 66.2 & 37.9 & 56.4 & 43.6 & 62.8 \\
VT-RDI & 68.8 & 78.7 & 98.2 & 82.5 & 27.9 & 54.5 & 56.1 & 32.8 & 45.8 & 37.9 & 58.3 \\
Admix-RDI & 74.2 & 80.7 & 98.7 & 86.8 & 20.9 & 59.4 & 56.1 & 26.7 & 42.7 & 34.1 & 58.0 \\
ODI & 78.3 & 77.1 & 97.6 & 87.0 & 43.8 & 67.3 & 70.0 & 49.5 & 65.9 & 55.4 & 69.2 \\
\rowcolor{Gray}CFM-RDI & \textbf{84.7} & \textbf{88.4} & 98.4 & \textbf{90.3} & \textbf{51.1} & \textbf{81.5} & \textbf{78.8} & \textbf{48.0} & \textbf{65.5} & \textbf{59.3} & \textbf{74.6}\\ \midrule
\textbf{\begin{tabular}[c]{@{}l@{}}Source : Inc-v3\end{tabular}} & \multicolumn{10}{c}{Target model} &  \\\cmidrule(l){2-11}
Attack & VGG-16 & RN-18 & RN-50 & DN-121 & Xcep & MB-v2 & EF-B0 & IR-v2 & Inc-v3 & Inc-v4 & Avg. \\ \midrule
DI & 2.9 & 2.4 & 3.4 & 5.0 & 1.9 & 1.8 & 3.7 & 3.0 & \textbf{99.2} & 4.2 & 12.8 \\
RDI & 3.5 & 3.8 & 4.0 & 7.0 & 3.1 & 3.0 & 5.9 & 6.3 & 98.7 & 7.1 & 14.2 \\
SI-RDI & 4.0 & 5.2 & 5.7 & 11.0 & 6.3 & 4.6 & 8.2 & 11.6 & 98.8 & 12.1 & 16.8 \\
VT-RDI & 5.9 & 8.9 & 9.4 & 13.2 & 7.4 & 5.9 & 9.8 & 12.3 & 98.7 & 14.7 & 18.6 \\
Admix-RDI & 6.3 & 6.5 & 8.8 & 12.8 & 6.0 & 6.1 & 10.9 & 12.2 & 98.7 & 13.6 & 18.2 \\
ODI & 14.3 & 14.9 & 16.7 & 32.3 & 20.3 & 13.7 & 25.3 & 26.4 & 95.6 & 31.6 & 29.1\\
\rowcolor{Gray} CFM-RDI & \textbf{22.9} & \textbf{26.8} & \textbf{26.2} & \textbf{39.1} & \textbf{34.1} & \textbf{27.1} & \textbf{38.6} & \textbf{36.2} & 95.9 & \textbf{44.8} & \textbf{39.2} \\
\bottomrule
\end{tabular}%
}
\caption{Targeted attack success rates (\%) against ten target models on the ImageNet-Compatible dataset.}
\label{tab:table1}
\vspace{-0.5cm}
\end{table*}

\noindent\textbf{Source and target models.}
We employed ten pre-trained neural networks as target networks: VGG-16 \cite{simonyan2014very}, ResNet-18 (RN-18) \cite{he2016deep}, ResNet-50 (RN-50) \cite{he2016deep}, DenseNet-121 (DN-121) \cite{huang2017densely}, Xception (Xcep) \cite{chollet2017xception}, MobileNet-v2 (MB-v2) \cite{sandler2018mobilenetv2}, EfficientNet-B0 (EF-B0) \cite{tan2019efficientnet}, Inception ResNet-v2 (IR-v2) \cite{szegedy2017inception}, Inception-v3 (Inc-v3) \cite{szegedy2016rethinking}, and Inception-v4 (Inc-v4) \cite{szegedy2017inception}. 
Additionally, we included an adversarially trained RN-50 network (adv-RN-50) \cite{wong2020fast}, which was trained with small $\ell_2$-norm-constrained adversarial examples ($||\bm{ \delta}||_2\leq0.1$), as it is effective in boosting the transfer success rate when used as a source model \cite{springer2021little}. 
We also added five Transformer-based classifiers: Vision Transformer (ViT) \cite{dosovitskiy2020image}, LeViT \cite{graham2021levit}, ConViT \cite{d2021convit}, Twins \cite{chu2021twins}, and Pooling-based Vision Transformer (PiT) \cite{heo2021rethinking}.
For the CIFAR-10 dataset, we used various ensemble models composed of three ResNet-20 \cite{he2016deep} networks (ens3-RN-20). They are trained under four defensive settings: standard training, ADP \cite{pang2019improving}, GAL \cite{kariyappa2019improving}, and DVERGE \cite{yang2020dverge}. 
The sources of the pre-trained model weights are described in Appendix.

\begin{table*}[!t]
\centering
\resizebox{0.75\textwidth}{!}{%
\begin{tabular}{lcccccccc}
 \toprule[0.15em]
\textbf{\begin{tabular}[c]{@{}l@{}}Source : RN-50\end{tabular}} & \multicolumn{6}{c}{Target model} & \multicolumn{1}{l}{} &  \\\cmidrule(l){2-7}
Attack & \begin{tabular}[c]{@{}c@{}}adv-\\ RN-50\end{tabular} & ViT & LeViT & ConViT & Twins & PiT & Avg. & \begin{tabular}[c]{@{}c@{}}Computation time\\ per image (sec)\end{tabular} \\\midrule
DI & 10.9  & 0.1 & 3.6 & 0.3 & 1.3 & 1.5 & 2.9 & 3.73 \\
RDI & 34.8  & 0.7 & 13.1 & 1.9 & 5.9 & 6.8 & 10.5 & 3.29 \\
SI-RDI & 59.9 & 2.9 & 29.4 & 6.3 & 15.5 & 17.9 & 22.0 & 16.16 \\
VT-RDI & 64.2 & 2.9 & 28.1 & 5.2 & 15.0 & 14.0 & 21.6 & 19.83 \\
Admix-RDI & 52.4 & 1.3 & 22.5 & 2.5 & 8.5 & 8.4 & 15.9 & 9.73 \\
ODI & 64.7 & \textbf{5.1} & 37.0 & \textbf{10.7} & 20.1 & \textbf{29.1} & 27.8 & 9.05 \\
\rowcolor{Gray}CFM-RDI & \textbf{75.5} & 4.3 & \textbf{46.1} & 8.9 & \textbf{25.2} & 24.7 & \textbf{30.8} & 3.72\\\midrule
\textbf{\begin{tabular}[c]{@{}l@{}}Source : adv-RN-50\end{tabular}} & \multicolumn{6}{c}{Target model} & \multicolumn{1}{l}{} &  \\\cmidrule(l){2-7}
Attack & \begin{tabular}[c]{@{}c@{}}adv-\\ RN-50\end{tabular} & ViT & LeViT & ConViT & Twins & PiT & Avg. & \begin{tabular}[c]{@{}c@{}}Computation time\\ per image (sec)\end{tabular} \\\midrule
DI & \textbf{98.9} & 5.7 & 36.9 & 10.1 & 19.2 & 20.5 & 31.9 & 3.77 \\
RDI & 98.8 & 10.8 & 49.5 & 19.9 & 29.4 & 35.8 & 40.7 & 3.29 \\
SI-RDI & 98.7& 19.4 & 57.6 & 35.3 & 35.2 & 52.1 & 49.7 & 16.34 \\
VT-RDI & 98.5& 10.6 & 46.3 & 20.0 & 27.1 & 34.4 & 39.5 & 19.83 \\
Admix-RDI & \textbf{98.9} & 12.1 & 55.5 & 23.1 & 32.4 & 38.9 & 43.5 & 9.86 \\
ODI & 97.3 & 22.2 & 57.7 & 38.8 & 40.0 & 54.9 & 51.8 & 9.04  \\
\rowcolor{Gray}CFM-RDI & 98.3 & \textbf{29.5} & \textbf{69.8} & \textbf{41.8} & \textbf{52.7} & \textbf{59.8} & \textbf{58.6} & 3.74\\
\bottomrule
\end{tabular}

}
\caption{Targeted attack success rates (\%) against a robust model and five Transformer-based classifiers with the ImageNet-Compatible dataset. We also report the average computation time to construct an adversarial example.}  \vspace{-0.3cm}
\label{tab:table2}
\end{table*}
\noindent\textbf{Baseline attacks.} 
We composed the baseline attacks using various combinations of eight existing techniques: DI \cite{xie2019improving}, RDI \cite{zou2020improving},  MI \cite{dong2018boosting}, TI \cite{dong2019evading}, SI \cite{lin2019nesterov}, VT \cite{wang2021enhancing}, Admix \cite{wang2021admix}, and ODI \cite{byun2022improving}.
We applied MI and TI techniques to all attack methods, so we omitted `MI-TI' when denoting them. Iteratively feeding fixed-size images can easily result in the overfitting of adversarial examples. Consequently, we opted for RDI as a common baseline technique in most cases. It is worth noting that, due to the computational intensity of ODI, we considered RDI as our primary baseline.
We followed \cite{byun2022improving} for the detailed setup of DI, RDI, TI, and ODI. Specifically, the scale multipliers of image sizes were $1 \sim\frac{330}{299}$ and $\frac{340}{299}$ for DI and RDI, respectively. We set the convolution kernel size for TI to $5\times5$, the transformation probability for DI to $0.7$, and the decay factor $\mu$ for MI to 1.0. For VT and SI, we set the number of samples and scales to 5, and $\beta$ of VT to $1.5$. For Admix, we set the mixing weight $w=0.2$ and the number of images to be mixed $N=3$ (\ie $m_2=3$ in \cite{wang2021admix}) following the experimental settings of \cite{wang2021admix}. 
The original Admix settings utilize SI with five scale copies in its internal loops. However, using SI in internal loops of Admix makes it difficult to directly compare the performance improvement of an image-level mixup in Admix and a feature-level mixup in CFM. For that reason, we basically set the number of scale copies of SI inside Admix to 1 (\ie $m_1=1$ in \cite{wang2021admix}). However, for comprehensive comparisons, we also included the results of Admix with the number of scale copies of 5 (\ie $m_1=5$ in \cite{wang2021admix}) in the Appendix. For both Admix and CFM, we used a batch size of 20 for fair comparisons.

\noindent \textbf{Settings for the CFM method.}
 We set the channel-wise mixing ratio $\bm{\alpha}$ to be randomly sampled from $\mathcal{U}(0, 0.75)$. We set the mixing probability $p$ to $0.1$ and $0.25$ for the ImageNet and CIFAR-10 datasets, respectively. Since the CIFAR-10 dataset has only ten classes, a larger value of $p$ is required to maximize the effectiveness of CFM. 
In addition, since the CFM method consumes one inference for storing clean features, we deducted the available remaining iterations to 299 for strictly fair comparisons. However, adversaries may also eliminate the need for this one additional inference for CFM by omitting the input transform at the first iteration.

\subsection{Experimental Results}
First, we conducted targeted attack experiments with the ImageNet-Compatible dataset. We used pre-trained RN-50 and Inc-v3 as source models and evaluated the targeted attack success rates on the ten target models. 

\noindent\textbf{Transfer success rates.}
Table \ref{tab:table1} shows the targeted attack success rates against the ten non-robust models of targeted adversarial examples generated from each source model. As shown in Table \ref{tab:table1}, CFM outperforms all baselines with a clear margin in all the source models. In particular, when the source model is Inc-v3, the performance improvement of CFM is remarkable, increasing the average attack success rate by more than 10\% over the second-best method. 

\begin{figure}[!t]
     \centering
         \includegraphics[width=0.47\textwidth,trim={0.9cm 0.0cm 0.1cm 0.0cm},clip]{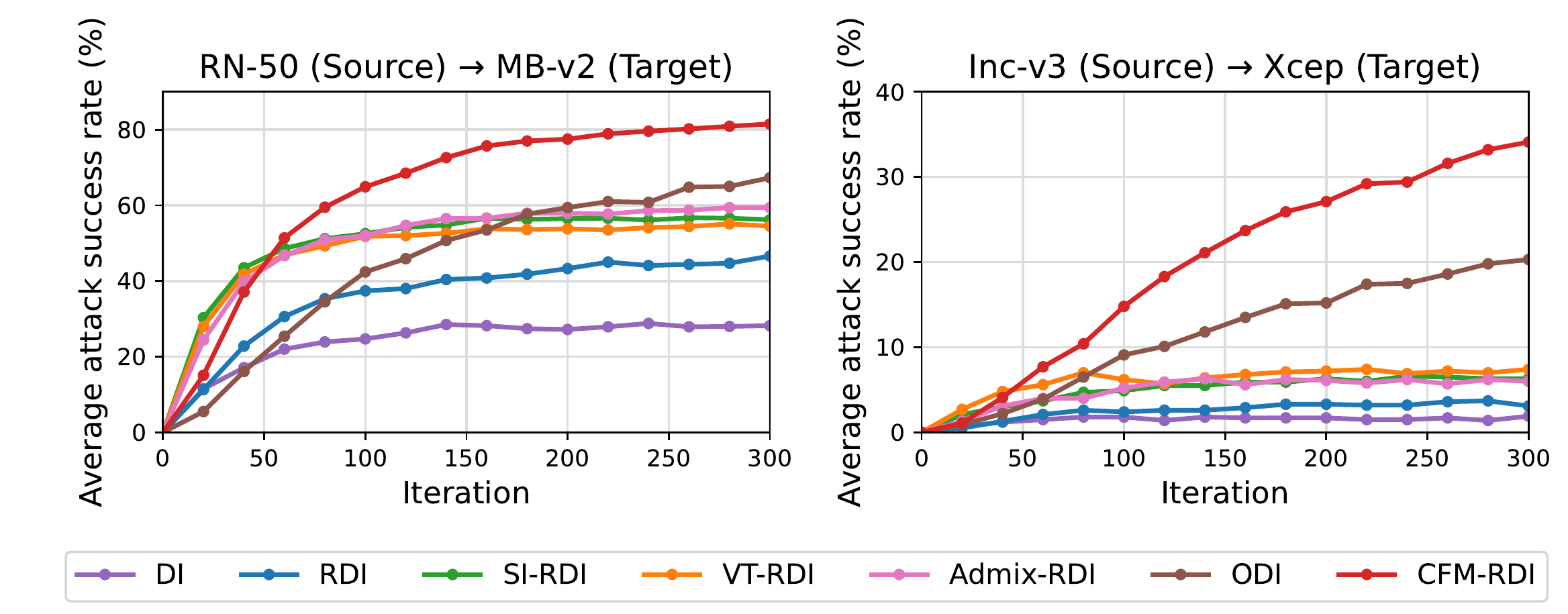}\vspace{-0.2cm}
         \caption{Targeted attack success rates (\%) based on the number of iterations. Best viewed in color.}\vspace{-0.4cm}
        \label{fig:fig3}
\end{figure}

Table \ref{tab:table2} shows the attack success rates with two source models, RN-50 and adv-RN-50, against an adversarially trained and five Transformer-based models. The proposed CFM technique overwhelms all the baseline techniques. In particular, the adversarial examples generated from adv-RN-50 record an average targeted attack success rate approaching 60\%.
Figure \ref{fig:fig3} shows the average attack success rate according to iterations for two cases. It can be observed that CFM outperforms other techniques in average attack success rates and takes longer to saturate.

More experimental results with different source models can be found in Appendix. We also provide visualizations of the generated adversarial examples in the Appendix for qualitative comparisons.

\noindent\textbf{Transfer-based attacks on the CIFAR-10 dataset.}
We also conducted transfer-based targeted attacks with the CIFAR-10 dataset. 
The primary purpose of this experiment is to evaluate the attack performance against four different ensembles of ResNet-20 models that were trained to be robust against transfer-based attacks. Table \ref{tab:table4} reports the transfer-attack success rates for five non-robust models along with the four ensemble models. Due to the small size of CIFAR-10 images, we could not apply ODI for this experiment, and we excluded the image size reduction at the end of RDI. The defensive model trained with DVERGE obviously lowers the average attack success rate of RDI to 14.9\%. Nevertheless, the CFM technique boosts the attack success rate for the DVERGE model from 14.9\% to 59.3\% and records an average attack success rate of 89.3\%.

\begin{table*}[!t]
\centering
\resizebox{\textwidth}{!}{%

\begin{tabular}{lccccccccccc}
\toprule[0.15em]
\textbf{\begin{tabular}[c]{@{}l@{}}Source : RN-50\end{tabular}} & \multicolumn{9}{c}{Target model} & \multicolumn{1}{l}{} &  \\\cmidrule(l){2-10}
\multirow{2}{*}{Attack} & \multirow{2}{*}{VGG-16} & \multirow{2}{*}{RN-18} & \multirow{2}{*}{MB-v2} & \multirow{2}{*}{Inc-v3} & \multirow{2}{*}{DN-121} & \multicolumn{4}{c}{ens3-RN-20} & \multirow{2}{*}{Avg.} & \multirow{2}{*}{\begin{tabular}[c]{@{}c@{}}Computation time\\ per image (sec)\end{tabular}} \\
 &  &  &  &  &  & Baseline & ADP & GAL & DVERGE &  &  \\ \midrule
DI & 66.4 & 71.5 & 62.7 & 71.1 & 84.2 & 77.9 & 56.5 & 14.3 & 15.6 & 57.8 & 0.64 \\
RDI & 66.4 & 70.9 & 64.1 & 73.4 & 82.8 & 76.3 & 55.8 & 13.5 & 14.9 & 57.6 & 0.59 \\
SI-RDI & 72.9 & 76.3 & 77.1 & 77.0 & 84.7 & 81.2 & 65.5 & 20.0 & 22.4 & 64.1 & 3.17 \\
VT-RDI & 89.8 & 87.1 & 92.6 & 92.9 & 93.7 & 94.4 & 82.3 & 24.3 & 31.3 & 76.5 & 3.82 \\
Admix-RDI & 74.2 & 78.8 & 76.2 & 82.7 & 89.2 & 85.2 & 66.4 & 17.3 & 18.4 & 65.4 & 1.98 \\
\rowcolor{Gray}CFM-RDI & \textbf{98.3} & \textbf{97.7} & \textbf{99.0} & \textbf{99.0} & \textbf{99.2} & \textbf{98.8} & \textbf{97.2} & \textbf{54.9} & \textbf{59.3} & \textbf{89.3} & 0.72
\\            \bottomrule
\end{tabular}%
}
\vspace{-0.2cm}
\caption{Targeted attack success rates (\%) against nine target models, including four ensemble-based defensive models on the CIFAR-10 dataset. We also evaluated the average computation time for crafting an adversarial example.}
\label{tab:table4}
\vspace{-0.1cm}
\end{table*}
\begin{table*}[!t]
\centering
\resizebox{\textwidth}{!}{
\begin{tabular}{cccccccccccccc}
\toprule[0.15em]
\multicolumn{3}{c}{Ablation} & \multicolumn{10}{c}{Target model} & \multicolumn{1}{l}{} \\\cmidrule(l){1-3} \cmidrule(l){4-13}
\begin{tabular}[c]{@{}c@{}}Mixing\\ clean features\end{tabular} & \begin{tabular}[c]{@{}c@{}}Channel-wise\\ mixing ratio\end{tabular} & Shuffle & Xcep & MB-v2 & EF-B0 & IR-v2 & Inc-v4 & ViT & LeViT & ConViT & Twins & PiT & Avg. \\ \midrule
\rowcolor{Gray}\checkmark & \checkmark & \checkmark & \textbf{67.1} & 82.4 & 83.4 & 64.7 & 67.4 & \textbf{29.5} & \textbf{69.8} & \textbf{41.8} & \textbf{52.7} & \textbf{59.8} & \textbf{61.9} \\
\checkmark & \checkmark &  & 57.8 & 76.1 & 77.8 & 58.2 & 60.1 & 23.4 & 64.9 & 38.9 & 46.6 & 52.8 & 55.7 \\
\checkmark &  & \checkmark & 65.0 & \textbf{83.1} & \textbf{83.6} & \textbf{65.3} & \textbf{68.1} & 26.6 & 69.4 & 40.5 & 50.6 & 57.5 & 61.0 \\
\checkmark &  &  & 58.1 & 76.3 & 78.7 & 59.5 & 61.2 & 23.9 & 63.6 & 39.2 & 48.5 & 54.0 & 56.3\\
 & \checkmark & \checkmark & 63.2 & 81.2 & 81.9 & 61.2 & 66.9 & 25.9 & 67.0 & 39.2 & 48.8 & 56.9 & 59.2 \\
 &  & \checkmark & 63.1 & 82.0 & 82.9 & 62.4 & 65.9 & 24.4 & 68.0 & 39.1 & 46.8 & 56.1 &59.1\\
             \bottomrule
\end{tabular}%
}
\vspace{-0.2cm}
\caption{Targeted attack success rates (\%) of CFM-RDI by ablating inner functions of the CFM modules. The source model is adv-RN-50.}
\vspace{-0.4cm}
\label{tab:table6}
\end{table*}

\noindent\textbf{Computational cost.}
In addition to the transfer success rates, computation time is also an important factor to consider, as it indicates the efficiency of a technique. To demonstrate the efficiency of CFM, we describe the average computation time for generating an adversarial example in the rightmost column of Table \ref{tab:table2} and Table \ref{tab:table4}. Since CFM modules add a marginal amount of computation, CFM-RDI increases only a small amount of computation time compared to other baselines. Note that each iterative attack was performed using a single NVIDIA Titan Xp GPU. 

\noindent\textbf{Combination with existing techniques.}
The CFM technique is compatible with many existing attack techniques. To demonstrate this, we attached Admix, SI, and VT to the CFM-RDI method. Due to space limitations, the experimental results are included in Appendix, but CFM showed further improved attack performance combined with other existing techniques.

\subsection{Ablation Study}
For an extensive ablation study, we investigated the range of mixing ratio $\bm {\alpha}$, mixing probability $p$, and the effect of internal functions of the CFM modules. For these ablation experiments, we carefully selected ten target models that are more difficult to disturb.

First, we evaluated how the transfer success rates vary by changing the values of the mixing probability $p$ and the upper bound of mixing ratios $\alpha_{max}$. In this experiment, we used adv-RN-50 as the source model and evaluated the transfer success rates on the ten target models, and detailed tabular experimental results can be seen in Appendix. CFM achieves the highest success rate when $p=0.1$ and $\alpha_{max}=0.75$, but it also achieves comparable attack success rates at other values. This indicates that CFM is not very sensitive to changes in hyperparameters and can consistently improve performance.

Next, we evaluated how the attack success rate varies by ablating each of the three internal functions of the CFM modules. Table \ref{tab:table6} shows the results of this ablation experiment. It can be seen that each internal function helps to improve the transferability of adversarial examples. Mixing clean features without shuffling means mixing only the clean features of the image itself, and even this improves the average attack success rate by more than 10\% compared to Admix (37.0\%), which mixes different images in the image domain. Without using the channel-wise mixing ratio, $\bm{\alpha}_i$ becomes a scalar (\ie $\alpha_i$) rather than a vector. Not mixing clean features means that each CFM module uses the features of other images being optimized in the batch without storing clean features. Since the features of the images in the batch are already perturbed by other targeted attacks, utilizing them for feature mixup degrades performance improvement, demonstrating the importance of mixing clean features.

Lastly, we also investigate the impact of batch size when applying CFM. We evaluated CFM-RDI with several batch sizes, but we could not observe significant differences. Specifically, the average success rates of CFM-RDI with batch sizes 5, 10, 20, and 30 over the ten target models in Table \ref{tab:table6} are 61.4\%, 61.6\%, 61.9\%, and 61.0\%, respectively.

\section{Conclusion}
In this paper, we proposed a novel approach to improve the transferability of targeted adversarial examples by introducing competition through the use of two types of competitor noises, which encourage the utilization of various features in attacks. Building upon this idea, we developed the Clean Feature Mixup (CFM) method, which efficiently simulates competitor noises in feature space by randomly mixing clean features of images in a batch. As CFM modules do not require extra backward passes, they require minimal computation, and this off-the-shelf model conversion-based method is easy to apply and compatible with many existing attacks.
Our extensive experiments on the ImageNet-Compatible and CIFAR-10 datasets demonstrate that CFM outperforms existing baselines by a significant margin, highlighting the effectiveness and versatility of our proposed method.
{\small
\bibliographystyle{ieee_fullname}
\bibliography{egbib}

\begin{thebibliography}{10}\itemsep=-1pt

\bibitem{byun2022improving}
Junyoung Byun, Seungju Cho, Myung-Joon Kwon, Hee-Seon Kim, and Changick Kim.
\newblock Improving the transferability of targeted adversarial examples
  through object-based diverse input.
\newblock In {\em Proceedings of the IEEE/CVF Conference on Computer Vision and
  Pattern Recognition}, pages 15244--15253, 2022.

\bibitem{chollet2017xception}
Fran{\c{c}}ois Chollet.
\newblock Xception: Deep learning with depthwise separable convolutions.
\newblock In {\em Proceedings of the IEEE conference on computer vision and
  pattern recognition}, pages 1251--1258, 2017.

\bibitem{chu2021twins}
Xiangxiang Chu, Zhi Tian, Yuqing Wang, Bo Zhang, Haibing Ren, Xiaolin Wei,
  Huaxia Xia, and Chunhua Shen.
\newblock Twins: Revisiting the design of spatial attention in vision
  transformers.
\newblock {\em Advances in Neural Information Processing Systems}, 34, 2021.

\bibitem{dong2018boosting}
Yinpeng Dong, Fangzhou Liao, Tianyu Pang, Hang Su, Jun Zhu, Xiaolin Hu, and
  Jianguo Li.
\newblock Boosting adversarial attacks with momentum.
\newblock In {\em Proceedings of the IEEE conference on computer vision and
  pattern recognition}, pages 9185--9193, 2018.

\bibitem{dong2019evading}
Yinpeng Dong, Tianyu Pang, Hang Su, and Jun Zhu.
\newblock Evading defenses to transferable adversarial examples by
  translation-invariant attacks.
\newblock In {\em Proceedings of the IEEE/CVF Conference on Computer Vision and
  Pattern Recognition}, pages 4312--4321, 2019.

\bibitem{dosovitskiy2020image}
Alexey Dosovitskiy, Lucas Beyer, Alexander Kolesnikov, Dirk Weissenborn,
  Xiaohua Zhai, Thomas Unterthiner, Mostafa Dehghani, Matthias Minderer, Georg
  Heigold, Sylvain Gelly, et~al.
\newblock An image is worth 16x16 words: Transformers for image recognition at
  scale.
\newblock {\em arXiv preprint arXiv:2010.11929}, 2020.

\bibitem{d2021convit}
St{\'e}phane d’Ascoli, Hugo Touvron, Matthew~L Leavitt, Ari~S Morcos, Giulio
  Biroli, and Levent Sagun.
\newblock Convit: Improving vision transformers with soft convolutional
  inductive biases.
\newblock In {\em International Conference on Machine Learning}, pages
  2286--2296. PMLR, 2021.

\bibitem{goodfellow2014explaining}
Ian~J Goodfellow, Jonathon Shlens, and Christian Szegedy.
\newblock Explaining and harnessing adversarial examples.
\newblock {\em arXiv preprint arXiv:1412.6572}, 2014.

\bibitem{graham2021levit}
Benjamin Graham, Alaaeldin El-Nouby, Hugo Touvron, Pierre Stock, Armand Joulin,
  Herv{\'e} J{\'e}gou, and Matthijs Douze.
\newblock Levit: a vision transformer in convnet's clothing for faster
  inference.
\newblock In {\em Proceedings of the IEEE/CVF International Conference on
  Computer Vision}, pages 12259--12269, 2021.

\bibitem{he2016deep}
Kaiming He, Xiangyu Zhang, Shaoqing Ren, and Jian Sun.
\newblock Deep residual learning for image recognition.
\newblock In {\em Proceedings of the IEEE conference on computer vision and
  pattern recognition}, pages 770--778, 2016.

\bibitem{heo2021rethinking}
Byeongho Heo, Sangdoo Yun, Dongyoon Han, Sanghyuk Chun, Junsuk Choe, and
  Seong~Joon Oh.
\newblock Rethinking spatial dimensions of vision transformers.
\newblock In {\em Proceedings of the IEEE/CVF International Conference on
  Computer Vision}, pages 11936--11945, 2021.

\bibitem{huang2017densely}
Gao Huang, Zhuang Liu, Laurens Van Der~Maaten, and Kilian~Q Weinberger.
\newblock Densely connected convolutional networks.
\newblock In {\em Proceedings of the IEEE conference on computer vision and
  pattern recognition}, pages 4700--4708, 2017.

\bibitem{huang2018intermediate}
Qian Huang, Zeqi Gu, Isay Katsman, Horace He, Pian Pawakapan, Zhiqiu Lin, Serge
  Belongie, and Ser-Nam Lim.
\newblock Intermediate level adversarial attack for enhanced transferability.
\newblock {\em arXiv preprint arXiv:1811.08458}, 2018.

\bibitem{kariyappa2019improving}
Sanjay Kariyappa and Moinuddin~K Qureshi.
\newblock Improving adversarial robustness of ensembles with diversity
  training.
\newblock {\em arXiv preprint arXiv:1901.09981}, 2019.

\bibitem{krizhevsky2009learning}
Alex Krizhevsky, Geoffrey Hinton, et~al.
\newblock Learning multiple layers of features from tiny images.
\newblock 2009.

\bibitem{kurakin2016adversarial}
Alexey Kurakin, Ian Goodfellow, Samy Bengio, et~al.
\newblock Adversarial examples in the physical world, 2016.

\bibitem{li2020towards}
Maosen Li, Cheng Deng, Tengjiao Li, Junchi Yan, Xinbo Gao, and Heng Huang.
\newblock Towards transferable targeted attack.
\newblock In {\em Proceedings of the IEEE/CVF Conference on Computer Vision and
  Pattern Recognition}, pages 641--649, 2020.

\bibitem{lin2019nesterov}
Jiadong Lin, Chuanbiao Song, Kun He, Liwei Wang, and John~E Hopcroft.
\newblock Nesterov accelerated gradient and scale invariance for adversarial
  attacks.
\newblock In {\em International Conference on Learning Representations}, 2019.

\bibitem{liu2016ssd}
Wei Liu, Dragomir Anguelov, Dumitru Erhan, Christian Szegedy, Scott Reed,
  Cheng-Yang Fu, and Alexander~C Berg.
\newblock Ssd: Single shot multibox detector.
\newblock In {\em European conference on computer vision}, pages 21--37.
  Springer, 2016.

\bibitem{madry2017towards}
Aleksander Madry, Aleksandar Makelov, Ludwig Schmidt, Dimitris Tsipras, and
  Adrian Vladu.
\newblock Towards deep learning models resistant to adversarial attacks.
\newblock {\em arXiv preprint arXiv:1706.06083}, 2017.

\bibitem{pang2019improving}
Tianyu Pang, Kun Xu, Chao Du, Ning Chen, and Jun Zhu.
\newblock Improving adversarial robustness via promoting ensemble diversity.
\newblock In {\em International Conference on Machine Learning}, pages
  4970--4979. PMLR, 2019.

\bibitem{redmon2016you}
Joseph Redmon, Santosh Divvala, Ross Girshick, and Ali Farhadi.
\newblock You only look once: Unified, real-time object detection.
\newblock In {\em Proceedings of the IEEE conference on computer vision and
  pattern recognition}, pages 779--788, 2016.

\bibitem{sandler2018mobilenetv2}
Mark Sandler, Andrew Howard, Menglong Zhu, Andrey Zhmoginov, and Liang-Chieh
  Chen.
\newblock Mobilenetv2: Inverted residuals and linear bottlenecks.
\newblock In {\em Proceedings of the IEEE conference on computer vision and
  pattern recognition}, pages 4510--4520, 2018.

\bibitem{simonyan2014very}
Karen Simonyan and Andrew Zisserman.
\newblock Very deep convolutional networks for large-scale image recognition.
\newblock {\em arXiv preprint arXiv:1409.1556}, 2014.

\bibitem{springer2021little}
Jacob Springer, Melanie Mitchell, and Garrett Kenyon.
\newblock A little robustness goes a long way: Leveraging robust features for
  targeted transfer attacks.
\newblock {\em Advances in Neural Information Processing Systems}, 34, 2021.

\bibitem{szegedy2017inception}
Christian Szegedy, Sergey Ioffe, Vincent Vanhoucke, and Alexander~A Alemi.
\newblock Inception-v4, inception-resnet and the impact of residual connections
  on learning.
\newblock In {\em Thirty-first AAAI conference on artificial intelligence},
  2017.

\bibitem{szegedy2016rethinking}
Christian Szegedy, Vincent Vanhoucke, Sergey Ioffe, Jon Shlens, and Zbigniew
  Wojna.
\newblock Rethinking the inception architecture for computer vision.
\newblock In {\em Proceedings of the IEEE conference on computer vision and
  pattern recognition}, pages 2818--2826, 2016.

\bibitem{tan2019efficientnet}
Mingxing Tan and Quoc Le.
\newblock Efficientnet: Rethinking model scaling for convolutional neural
  networks.
\newblock In {\em International conference on machine learning}, pages
  6105--6114. PMLR, 2019.

\bibitem{tramer2018ensemble}
Florian Tram{\`e}r, Alexey Kurakin, Nicolas Papernot, Ian Goodfellow, Dan
  Boneh, and Patrick McDaniel.
\newblock Ensemble adversarial training: Attacks and defenses.
\newblock In {\em International Conference on Learning Representations}, 2018.

\bibitem{wang2021enhancing}
Xiaosen Wang and Kun He.
\newblock Enhancing the transferability of adversarial attacks through variance
  tuning.
\newblock In {\em Proceedings of the IEEE/CVF Conference on Computer Vision and
  Pattern Recognition}, pages 1924--1933, 2021.

\bibitem{wang2021admix}
Xiaosen Wang, Xuanran He, Jingdong Wang, and Kun He.
\newblock Admix: Enhancing the transferability of adversarial attacks.
\newblock In {\em Proceedings of the IEEE/CVF International Conference on
  Computer Vision}, pages 16158--16167, 2021.

\bibitem{wang2021feature}
Zhibo Wang, Hengchang Guo, Zhifei Zhang, Wenxin Liu, Zhan Qin, and Kui Ren.
\newblock Feature importance-aware transferable adversarial attacks.
\newblock In {\em Proceedings of the IEEE/CVF International Conference on
  Computer Vision}, pages 7639--7648, 2021.

\bibitem{wong2020fast}
Eric Wong, Leslie Rice, and J~Zico Kolter.
\newblock Fast is better than free: Revisiting adversarial training.
\newblock {\em arXiv preprint arXiv:2001.03994}, 2020.

\bibitem{xie2019improving}
Cihang Xie, Zhishuai Zhang, Yuyin Zhou, Song Bai, Jianyu Wang, Zhou Ren, and
  Alan~L Yuille.
\newblock Improving transferability of adversarial examples with input
  diversity.
\newblock In {\em Proceedings of the IEEE/CVF Conference on Computer Vision and
  Pattern Recognition}, pages 2730--2739, 2019.

\bibitem{yang2020dverge}
Huanrui Yang, Jingyang Zhang, Hongliang Dong, Nathan Inkawhich, Andrew Gardner,
  Andrew Touchet, Wesley Wilkes, Heath Berry, and Hai Li.
\newblock Dverge: diversifying vulnerabilities for enhanced robust generation
  of ensembles.
\newblock {\em Advances in Neural Information Processing Systems},
  33:5505--5515, 2020.

\bibitem{zhang2018mixup}
Hongyi Zhang, Moustapha Cisse, Yann~N Dauphin, and David Lopez-Paz.
\newblock mixup: Beyond empirical risk minimization.
\newblock In {\em International Conference on Learning Representations}, 2018.

\bibitem{zhao2021success}
Zhengyu Zhao, Zhuoran Liu, and Martha Larson.
\newblock On success and simplicity: A second look at transferable targeted
  attacks.
\newblock {\em Advances in Neural Information Processing Systems}, 34, 2021.

\bibitem{zou2020improving}
Junhua Zou, Zhisong Pan, Junyang Qiu, Xin Liu, Ting Rui, and Wei Li.
\newblock Improving the transferability of adversarial examples with
  resized-diverse-inputs, diversity-ensemble and region fitting.
\newblock In {\em European Conference on Computer Vision}, pages 563--579.
  Springer, 2020.

\end{thebibliography}
}

\end{document}


\title{Appendix for Introducing Competition to Boost the Transferability of \\Targeted Adversarial Examples through Clean Feature Mixup\vspace{-0.5cm}}

\maketitle

\renewcommand{\thesection}{\thesection\Alph{section}}
\appendix
\addcontentsline{toc}{section}{Appendices}

\section{Algorithm} 
The CFM method is compatible with many existing attack methods, and as an example, the pseudo-codes of the CFM-RDI-MI-TI method are described in Algorithm \ref{alg:cfm}.

\begin{algorithm}[!t]
    \algnewcommand\algorithmicinput{\textbf{Input:}}
    \algnewcommand\Input{\item[\algorithmicinput]}
    \algnewcommand\algorithmicoutput{\textbf{Output:}}
    \algnewcommand\Output{\item[\algorithmicoutput]}
    \caption{CFM-RDI-MI-TI}
    \label{alg:cfm}
    \begin{flushleft}
    \textbf{Input:} A classifier $f$; a clean example $\vct x$; a target label $y_{t}$. \\
    \textbf{Input:} Adversary's objective $\mathcal{L}$; the maximum iterations $T$; $\ell_\infty$ perturbation bounds $\epsilon$; step size $\eta$; decay factor $\mu$; Gaussian kernel $\mW$ for TI. \\
    \textbf{Input:} mixing probabilty $p$; upper bounds for mixing ratios $\alpha_{max}$ for CFM modules.
    \\
    \textbf{Output:} An adversarial example $\vct x^{adv}$
\end{flushleft}

	\begin{algorithmic}[1]
	    \State $f'=AttachCFM(f; p, \alpha_{max})$ \Comment{Attach CFM modules to \textit{conv} and \textit{fc} layers}
		\State Store clean features into CFM modules via $f'(\vct x)$ 
		\State $\vct g_1 = 0$; $\vct x_1^{adv}=\vct x$
		\For{$t = 1 \rightarrow T-1$} 
		    \State Compute the gradients with RDI input transforms via $f'$
		    \begin{equation}
		        \hat{\vct g}_{t+1} = \bigtriangledown_{\vct x^{adv}_t}\mathcal{L}(f'(RDI(\vct x^{adv}_t)), y_{t})
		    \end{equation}
		   \State $\tilde{\vct g}_{t+1} = \mu \cdot \vct g_t + \frac{\hat{\vct g}_{t+1}}{\|\hat{\vct g}_{t+1}\|_1}$ \Comment{Apply MI}
		  \State $\vct g_{t+1}=\mW \ast \tilde{\vct g}_{t+1}$ \Comment{Apply TI}
		    \State $\vct x_{t+1}^{adv} = \vct x_t^{adv} - \eta \cdot \text{sign}(\vct g_{t+1})$ \Comment{Apply FGSM}
		     \State $\vct x_{t+1}^{adv}=Clip_{\vct x}^{\epsilon}(\vct x_{t+1}^{adv})$
		\EndFor
		\State $\vct x^{adv}=\vct x_T^{adv}$
        \State \Return $\vct x^{adv}$
	\end{algorithmic} 
\end{algorithm}

\section{References to Pre-trained Models} 
\subsection{Pre-trained Models on the ImageNet Dataset}
We used a total of 16 models, and the sources of the pre-trained weights of the models are as follows.

The weights for the following six models are downloaded from TorchVision library\footnote{\url{https://github.com/pytorch/vision}}: VGG-16 \cite{simonyan2014very}, ResNet-18 (RN-18) \cite{he2016deep}, ResNet-50 (RN-50) \cite{he2016deep}, DenseNet-121 (DN-121) \cite{huang2017densely}, MobileNet-v2 (MB-v2) \cite{sandler2018mobilenetv2},
 Inception-v3 (Inc-v3) \cite{szegedy2016rethinking}. 
 
 The weights for the following nine models are downloaded from Pytorch Image Models (timm) library \cite{rw2019timm}:
 Xception (Xcep) \cite{chollet2017xception}, EfficientNet-B0 (EF-B0) \cite{tan2019efficientnet}, Inception ResNet-v2 (IR-v2) \cite{szegedy2017inception}, Inception-v4 (Inc-v4) \cite{szegedy2017inception}, Vision Transformer (ViT) \cite{dosovitskiy2020image}, LeViT \cite{graham2021levit}, ConViT \cite{d2021convit}, Twins \cite{chu2021twins}, and Pooling-based Vision Transformer (PiT) \cite{heo2021rethinking}. The pre-trained weights for the adversarially trained RN-50 (adv-RN-50) \cite{wong2020fast} is provided by the official repository of \cite{salman2020adversarially}. 
 
 The adv-RN-50 is adversarially trained on small $\ell_2$-norm-constrained adversarial examples ($||\bm{ \delta}||_2\leq0.1$), which is recently demonstrated to be effective in boosting the transfer success rate when used as a source model \cite{springer2021little}.
 
\subsection{Pre-trained Models on the CIFAR-10 Dataset}
The pre-trained weights for the following six models are provided by \cite{huy_phan_2021_4431043}:
VGG-16 \cite{simonyan2014very}, ResNet-18 (RN-18) \cite{he2016deep}, ResNet-50 (RN-50) \cite{he2016deep}, DenseNet-121 (DN-121) \cite{huang2017densely}, MobileNet-v2 (MB-v2) \cite{sandler2018mobilenetv2}, and Inception-v3 (Inc-v3) \cite{szegedy2016rethinking}.
 
We used four ensemble models composed of three ResNet-20 \cite{he2016deep} networks (ens3-RN-20). They are trained under four defensive settings: standard training, ADP \cite{pang2019improving}, GAL \cite{kariyappa2019improving}, and DVERGE \cite{yang2020dverge}. The pre-trained weights for the four ensemble models are provided by \cite{yang2020dverge}.

\section{Additional Experimental Results}
\subsection{Visualization of Generated Adversarial Examples}
\begin{figure*}[!t]
     \centering
         \centering
         \includegraphics[width=\textwidth,trim={0cm 0.0cm 0cm 0.0cm},clip]{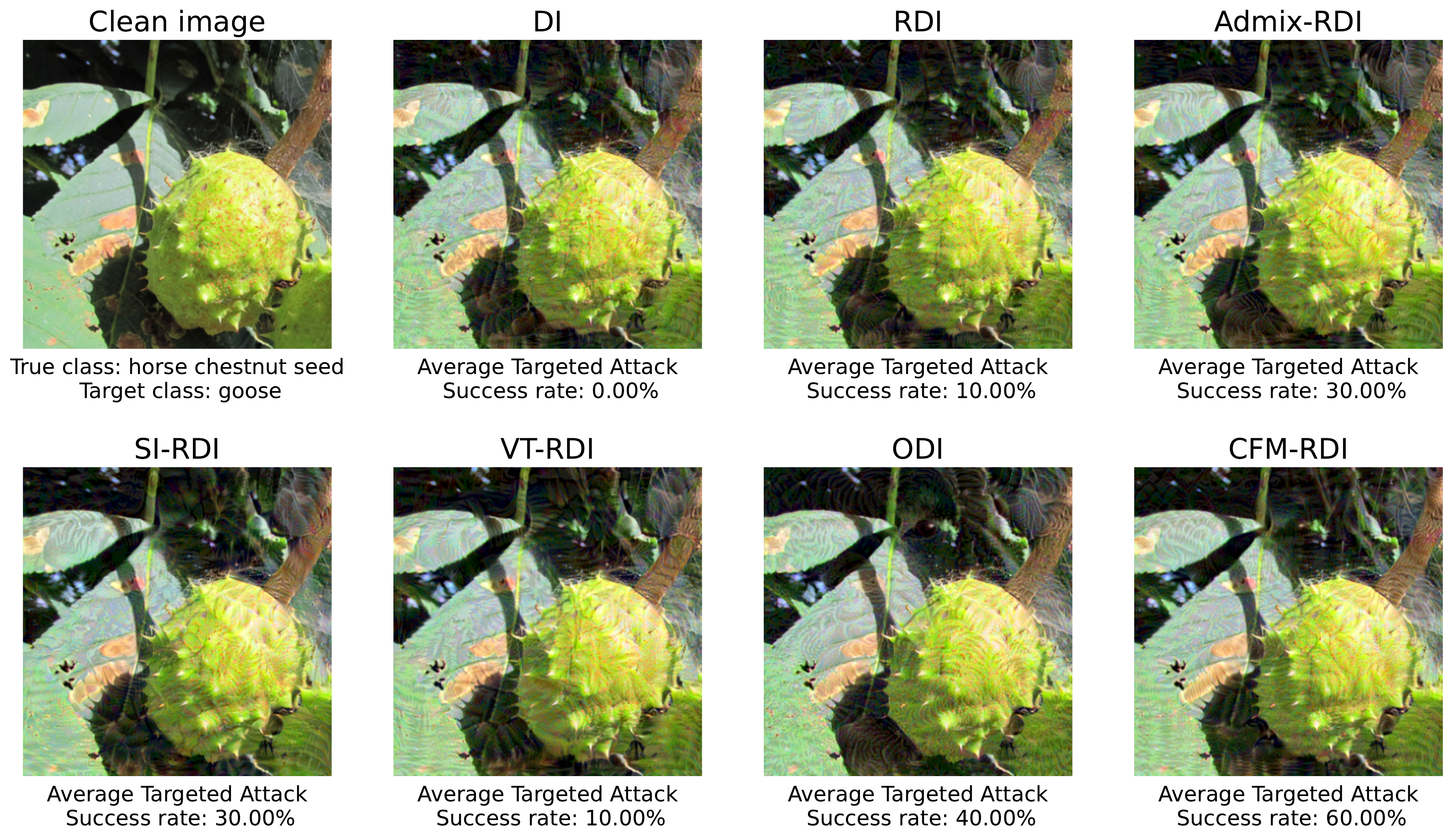}
              \includegraphics[width=\textwidth,trim={0cm 0.0cm 0cm 0.0cm},clip]{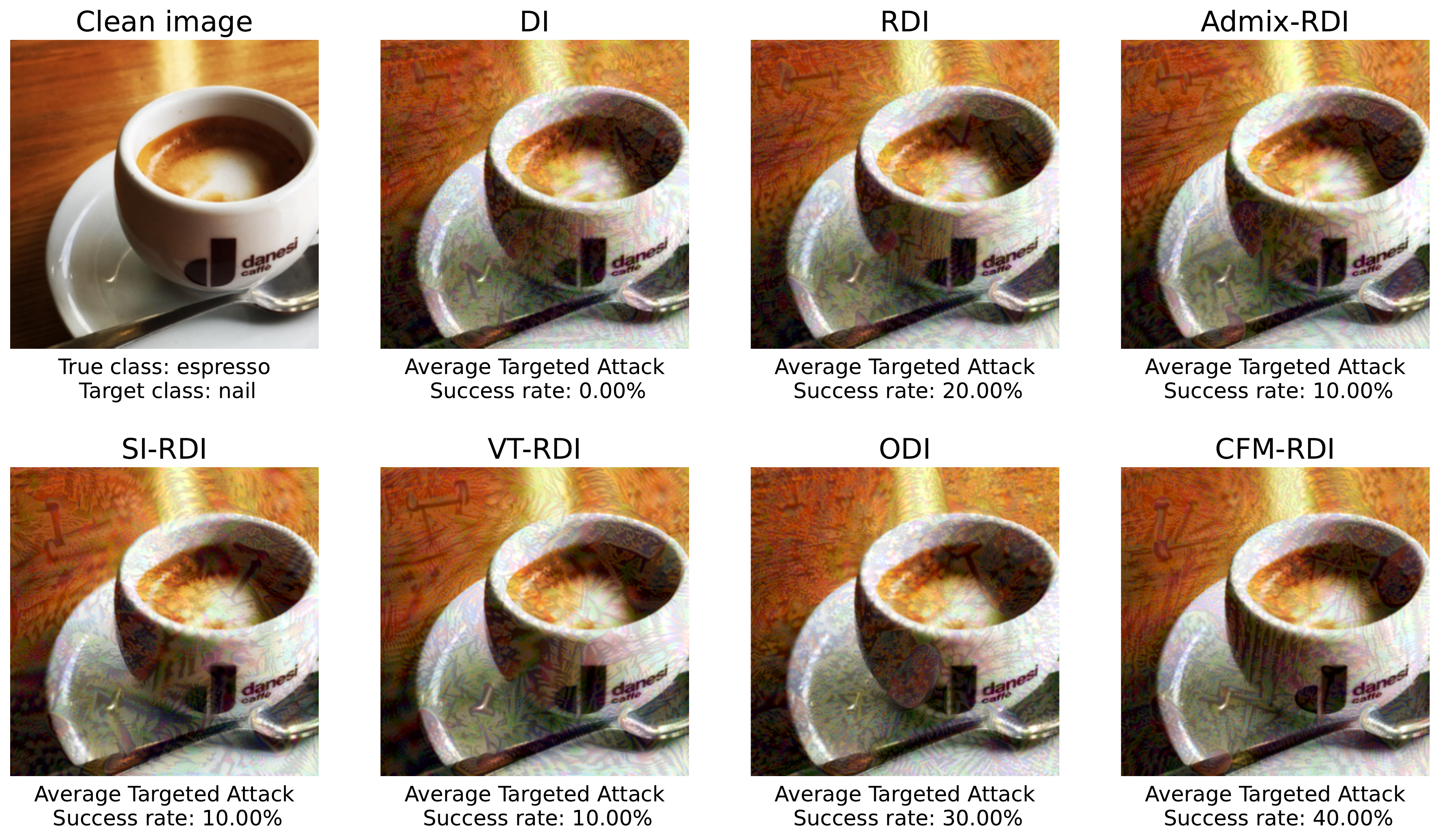}
         \caption{Visualization of generated adversarial examples. The source model is RN-50. 
Each average targeted attack success rate was calculated over the ten carefully selected target models, which are more difficult to confuse. For example, an average targeted attack success rate of 50\% means that 5 out of 10 target models recognize the adversarial example as the target class.}
        \label{fig:fig1}
\end{figure*}

\begin{figure*}[!t]
     \centering
         \centering
         \includegraphics[width=\textwidth,trim={0cm 0.0cm 0cm 0.0cm},clip]{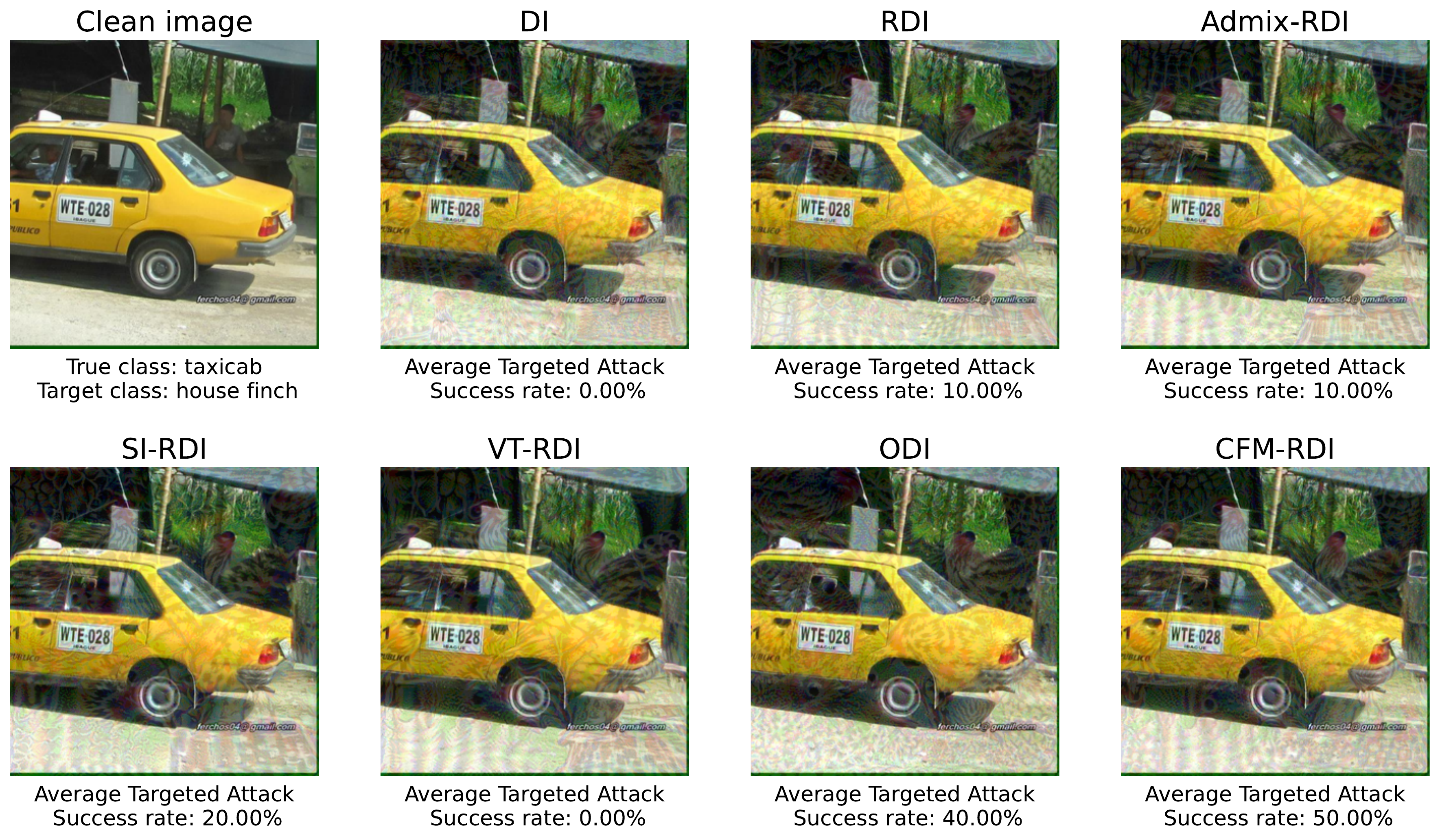}
              \includegraphics[width=\textwidth,trim={0cm 0.0cm 0cm 0.0cm},clip]{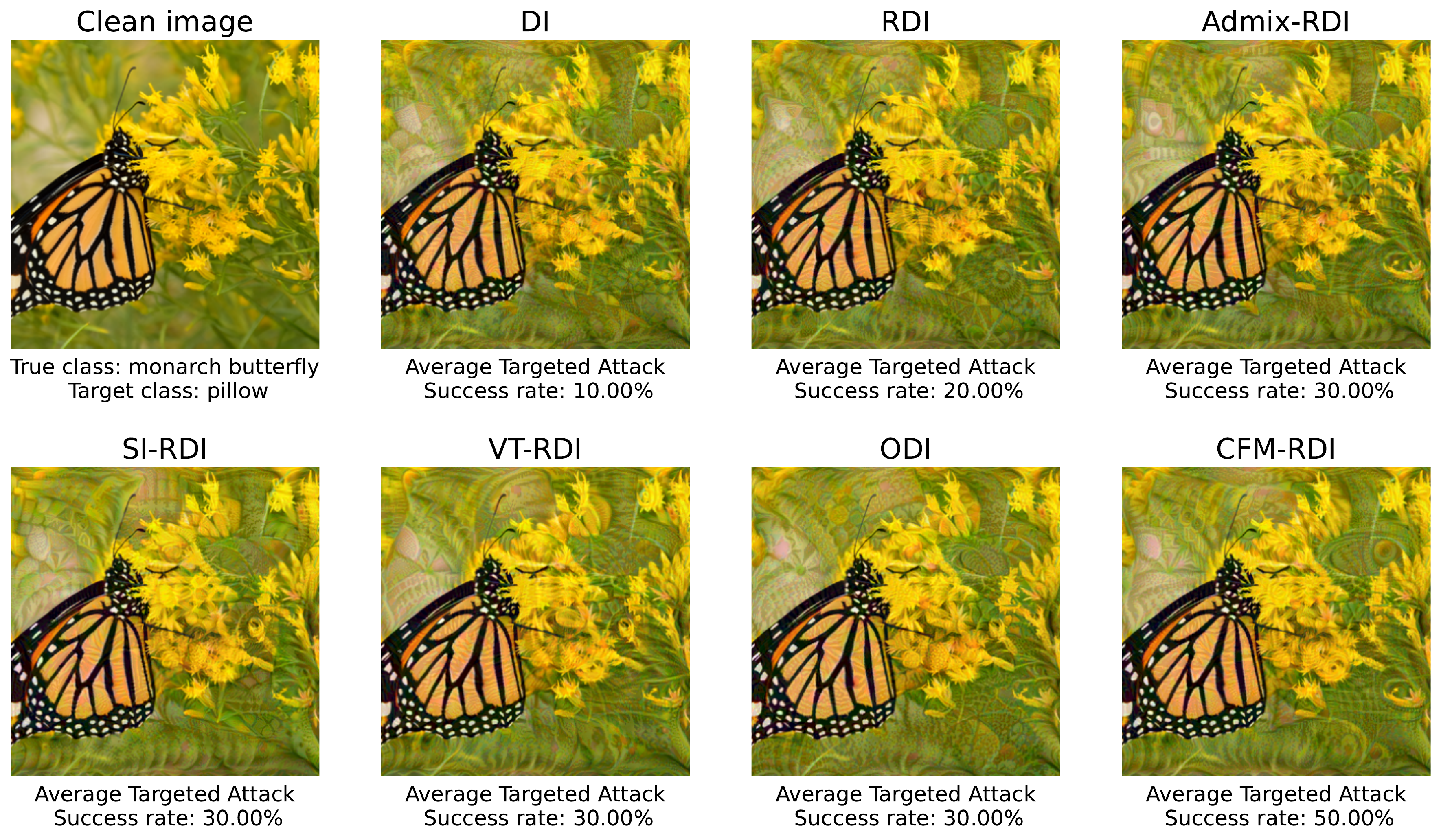}
         \caption{Visualization of generated adversarial examples. The source model is RN-50. 
Each average targeted attack success rate was calculated over the ten carefully selected target models, which are more difficult to confuse. For example, an average targeted attack success rate of 50\% means that 5 out of 10 target models recognize the adversarial example as the target class.}
        \label{fig:fig2}
\end{figure*}

\begin{figure*}[!t]
     \centering
         \centering
         \includegraphics[width=\textwidth,trim={0cm 0.0cm 0cm 0.0cm},clip]{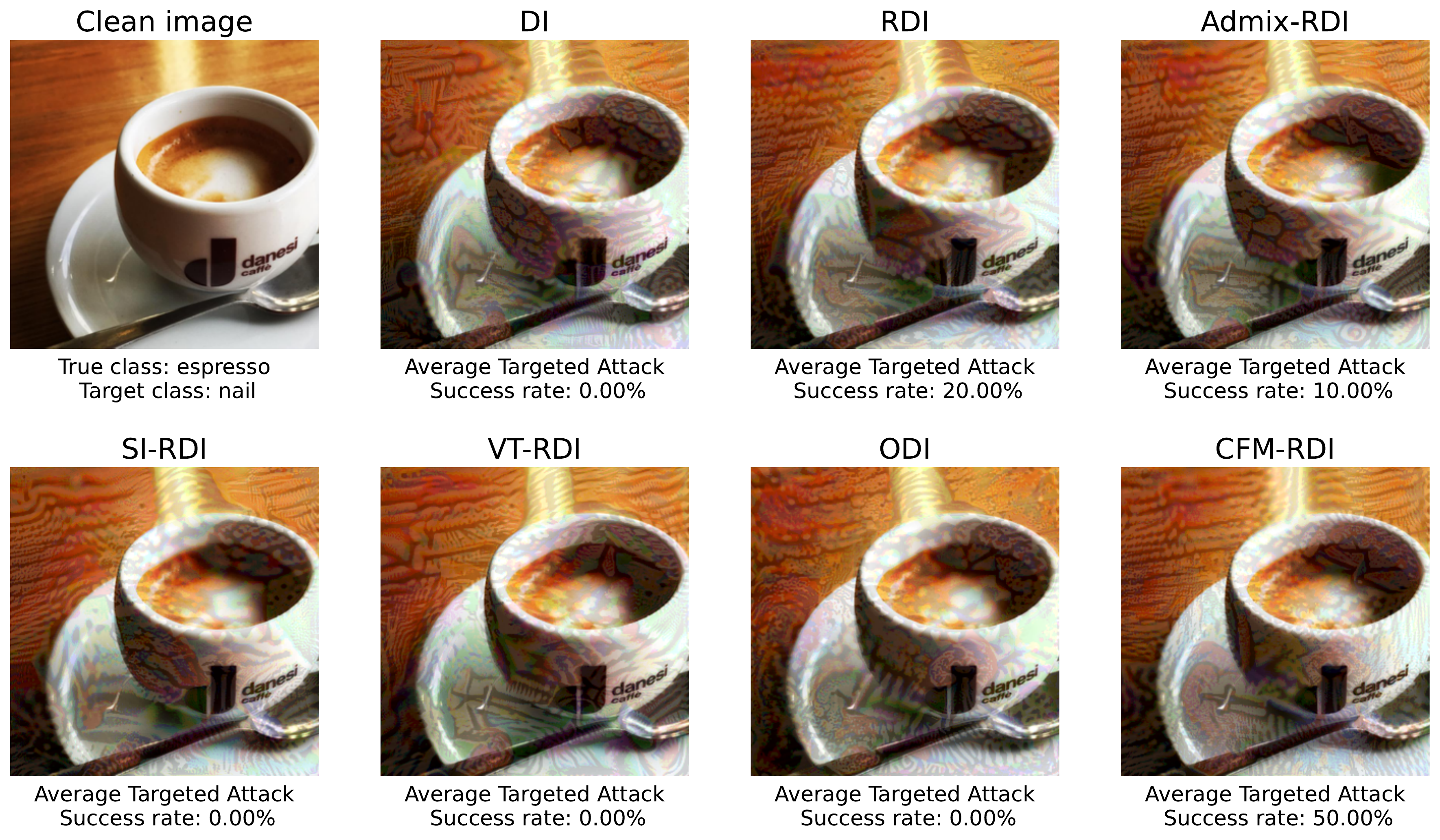}
              \includegraphics[width=\textwidth,trim={0cm 0.0cm 0cm 0.0cm},clip]{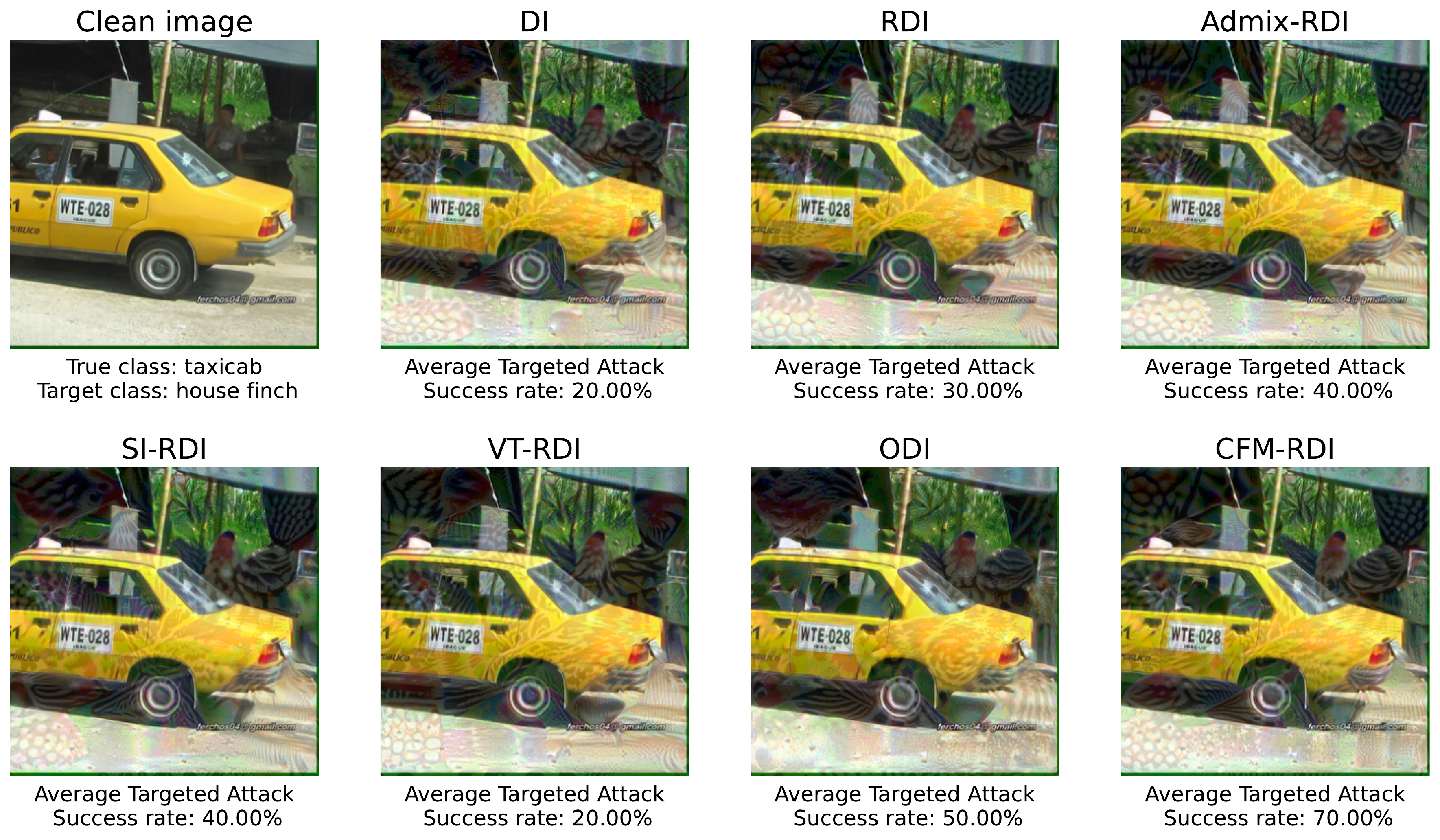}
         \caption{Visualization of generated adversarial examples. The source model is adv-RN-50. 
Each average targeted attack success rate was calculated over the ten carefully selected target models, which are more difficult to confuse. For example, an average targeted attack success rate of 50\% means that 5 out of 10 target models recognize the adversarial example as the target class.}
        \label{fig:fig3}
\end{figure*}

\begin{figure*}[!t]
     \centering
         \centering
         \includegraphics[width=\textwidth,trim={0cm 0.0cm 0cm 0.0cm},clip]{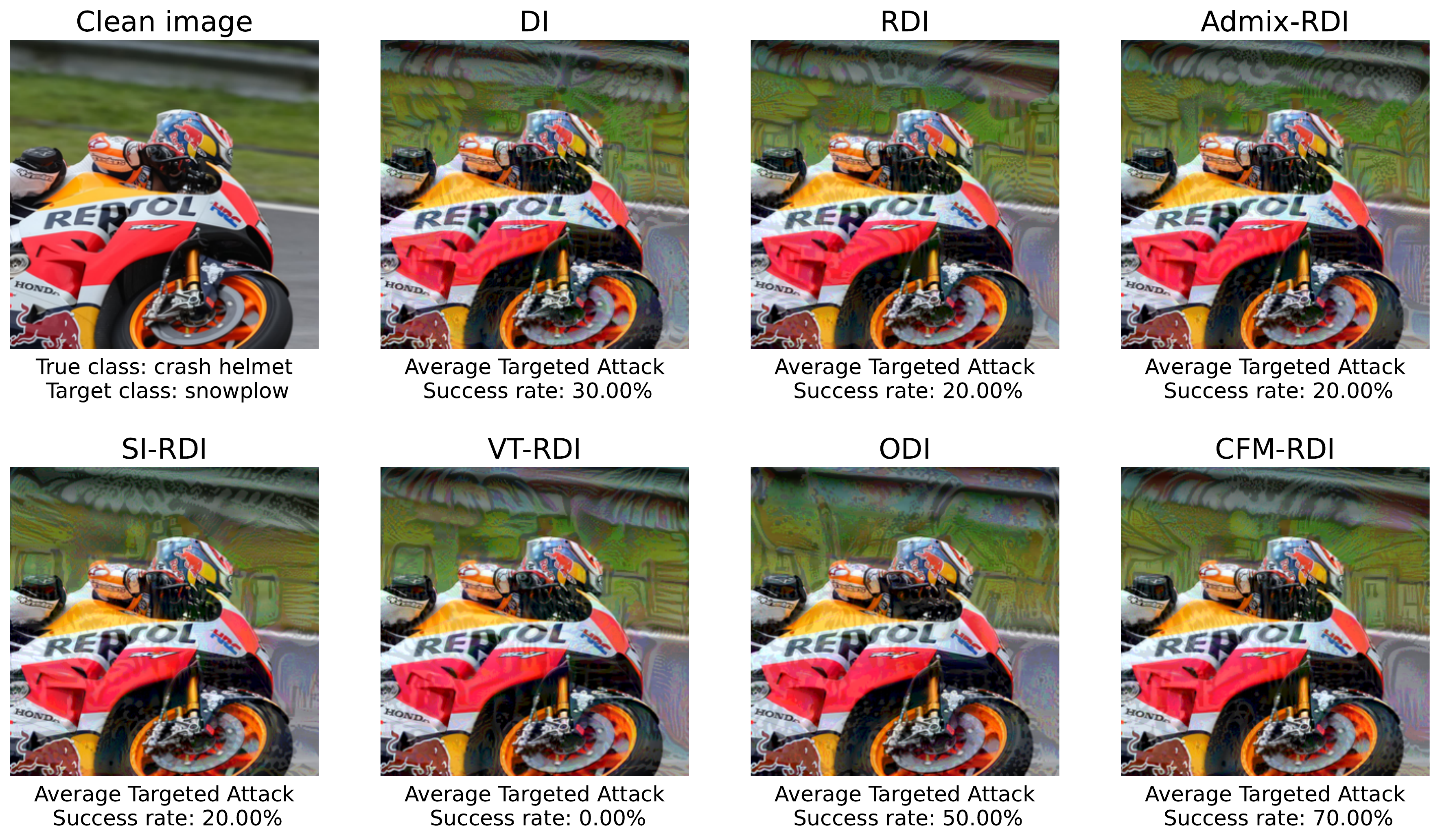}
              \includegraphics[width=\textwidth,trim={0cm 0.0cm 0cm 0.0cm},clip]{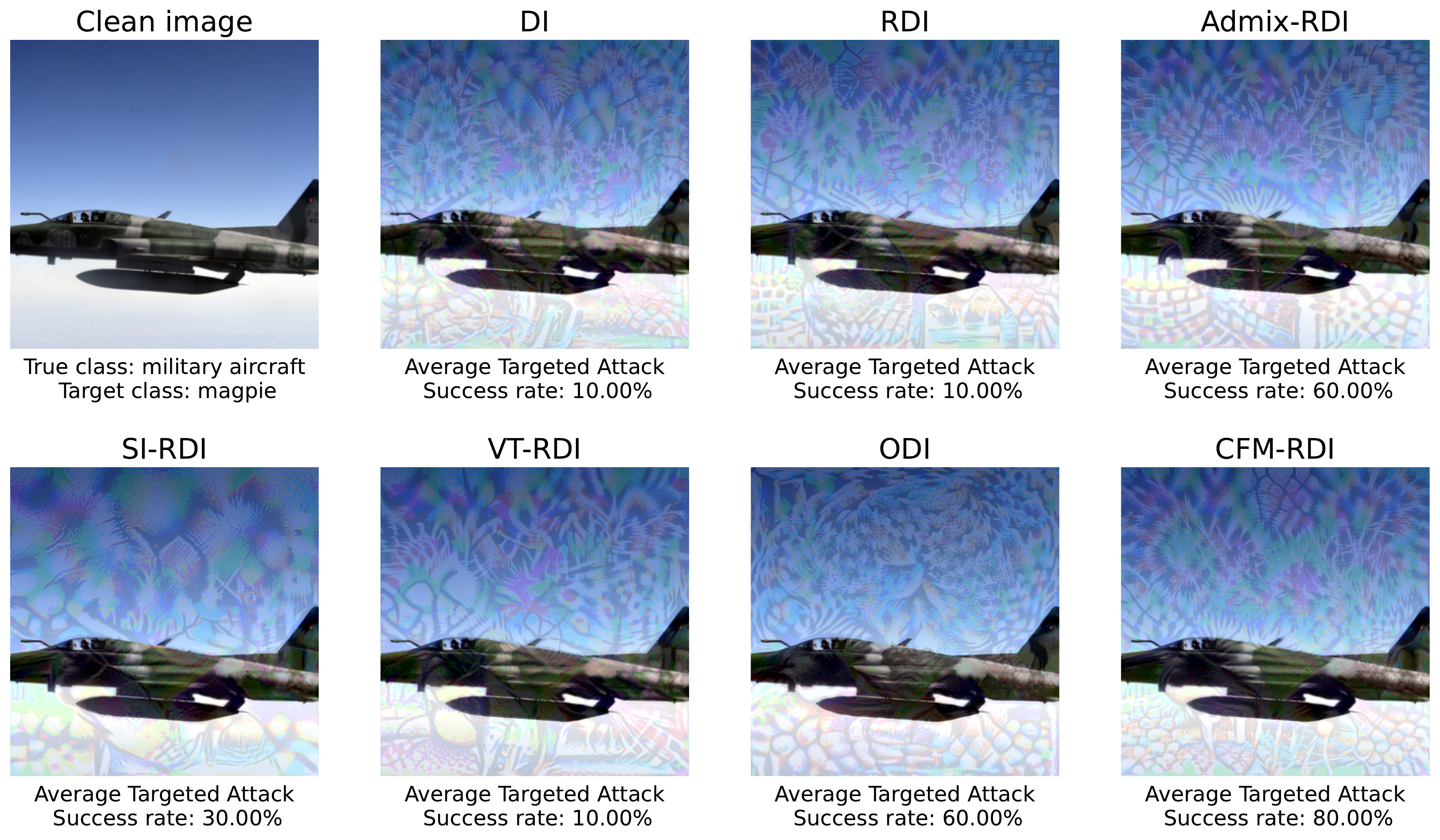}
         \caption{Visualization of generated adversarial examples. The source model is adv-RN-50. 
Each average targeted attack success rate was calculated over the ten carefully selected target models, which are more difficult to confuse. For example, an average targeted attack success rate of 50\% means that 5 out of 10 target models recognize the adversarial example as the target class.}
        \label{fig:fig4}
\end{figure*}

\begin{figure*}[!t]
     \centering
         \centering
         \includegraphics[width=\textwidth,trim={0cm 0.0cm 0cm 0.0cm},clip]{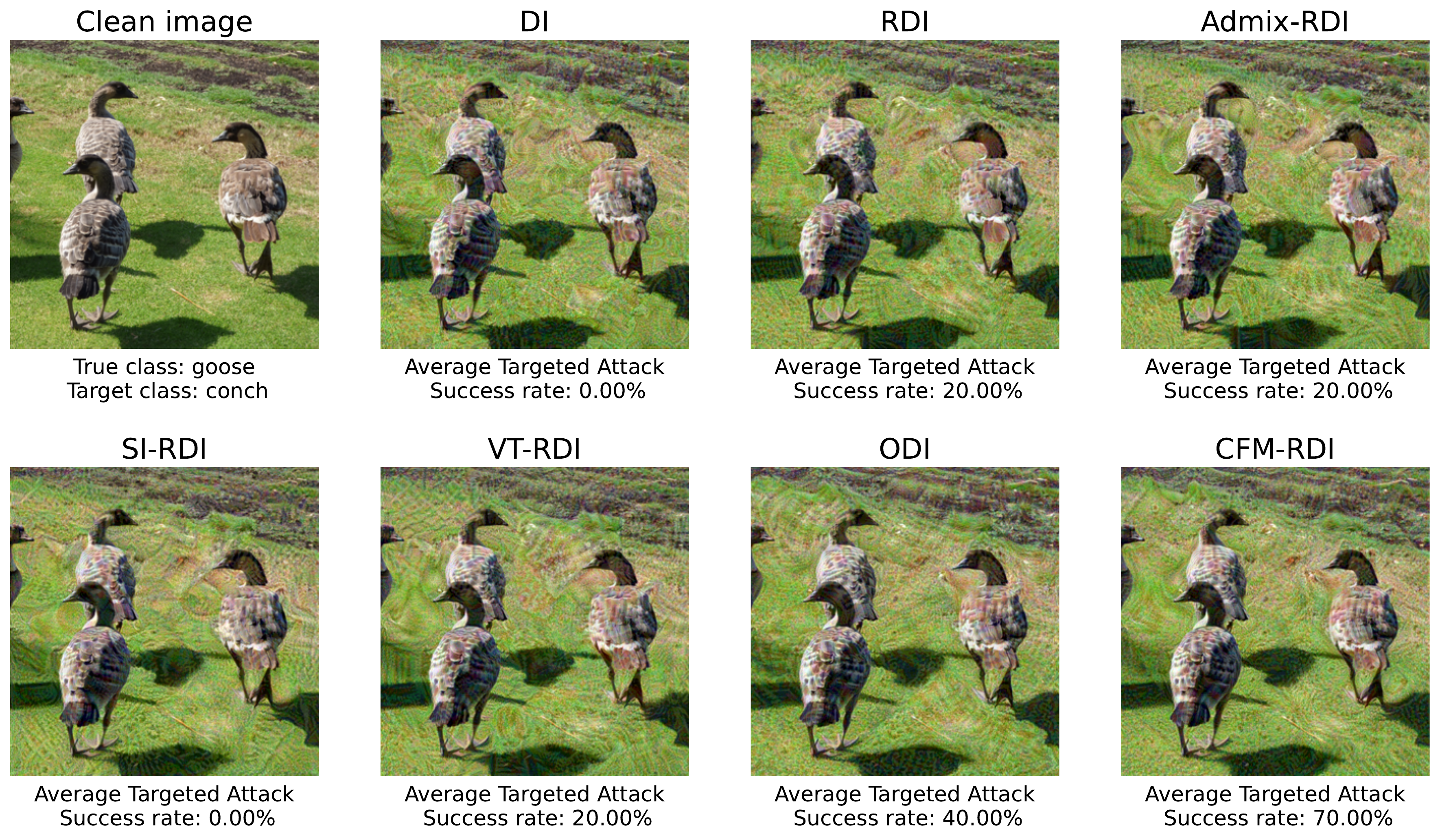}
              \includegraphics[width=\textwidth,trim={0cm 0.0cm 0cm 0.0cm},clip]{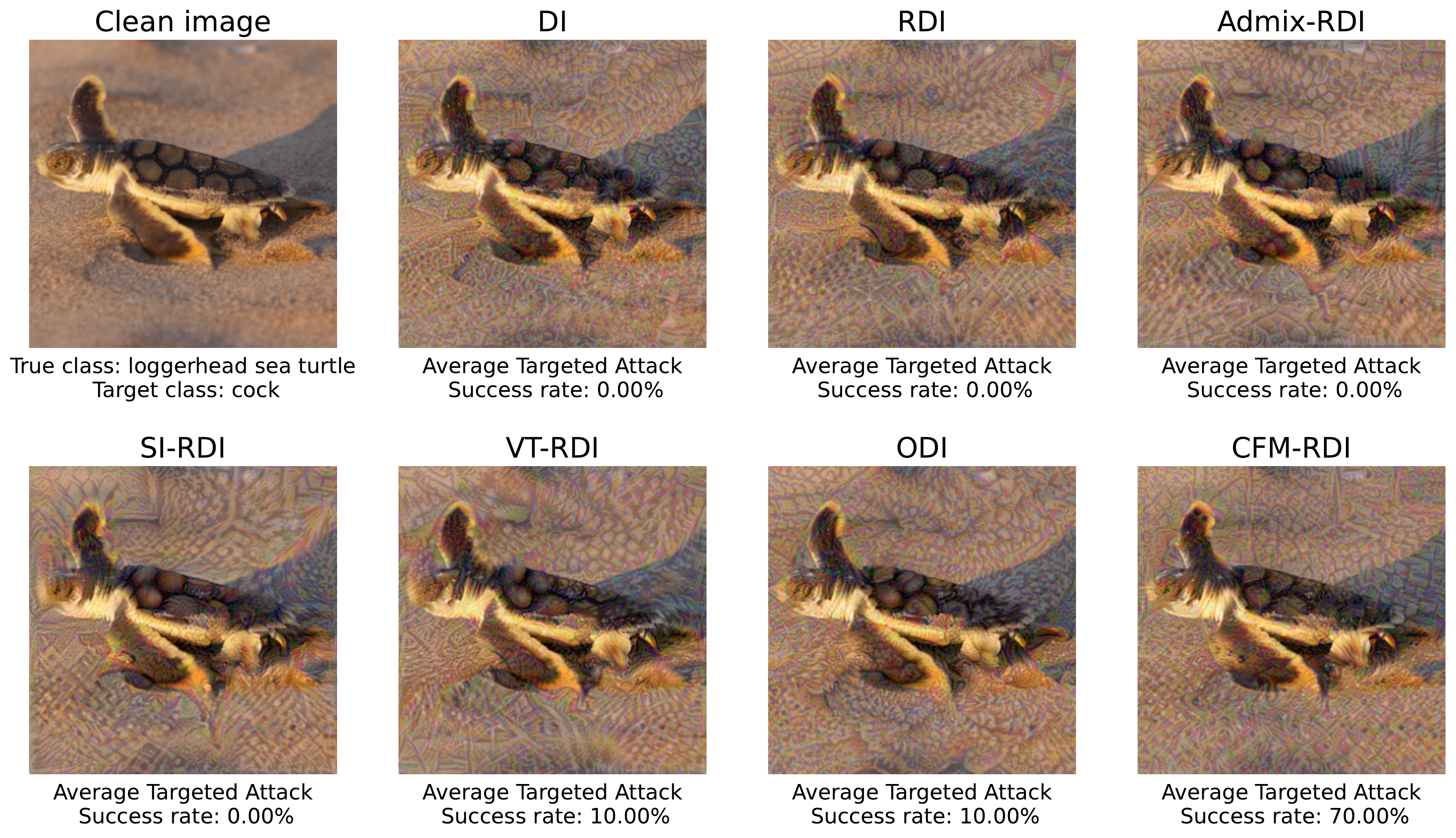}
         \caption{Visualization of generated adversarial examples. The source model is Inc-v3. 
Each average targeted attack success rate was calculated over the ten carefully selected target models, which are more difficult to confuse. For example, an average targeted attack success rate of 50\% means that 5 out of 10 target models recognize the adversarial example as the target class.}
        \label{fig:fig5}
\end{figure*}

\begin{figure*}[!t]
     \centering
         \centering
         \includegraphics[width=\textwidth,trim={0cm 0.0cm 0cm 0.0cm},clip]{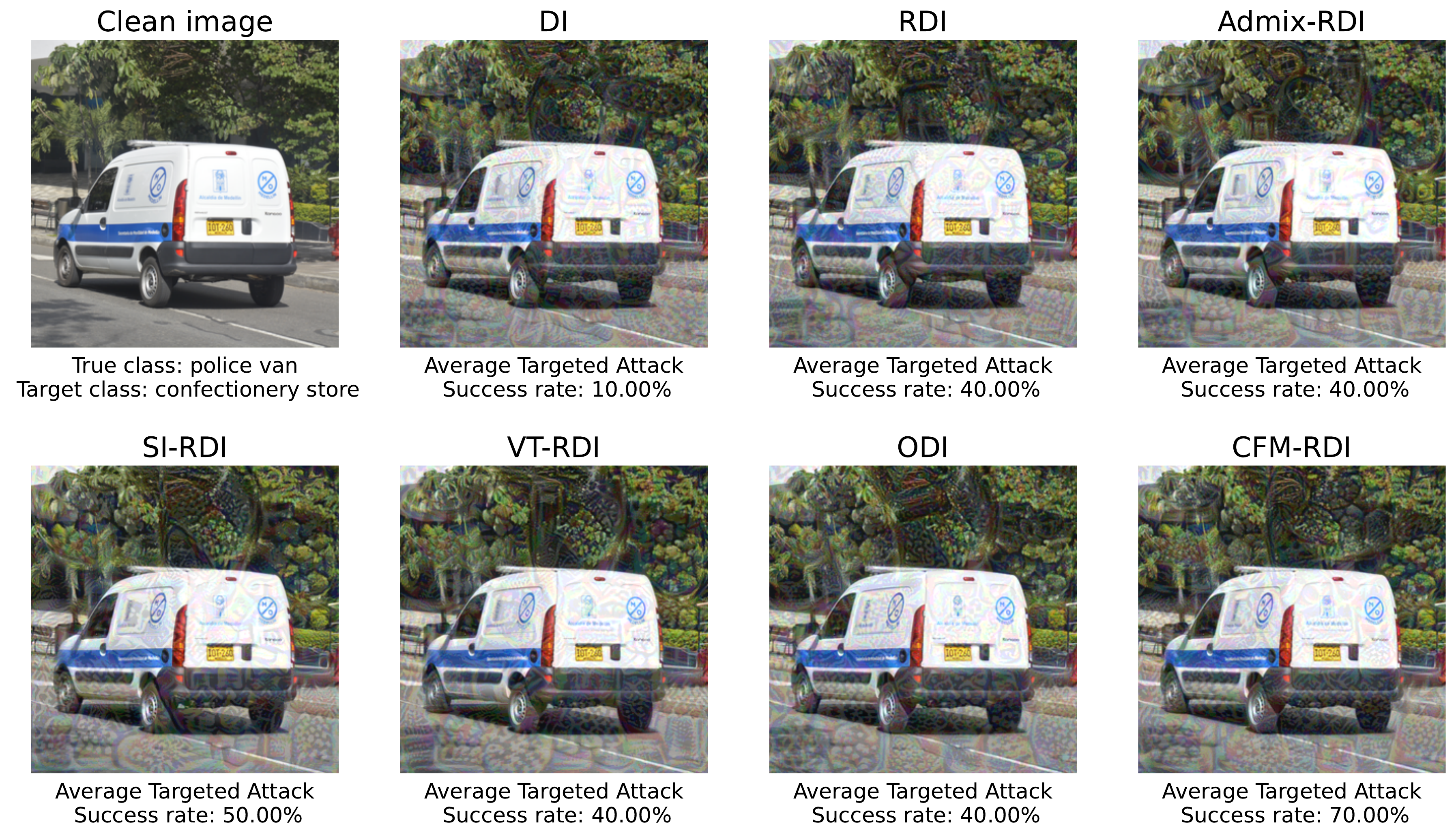}
              \includegraphics[width=\textwidth,trim={0cm 0.0cm 0cm 0.0cm},clip]{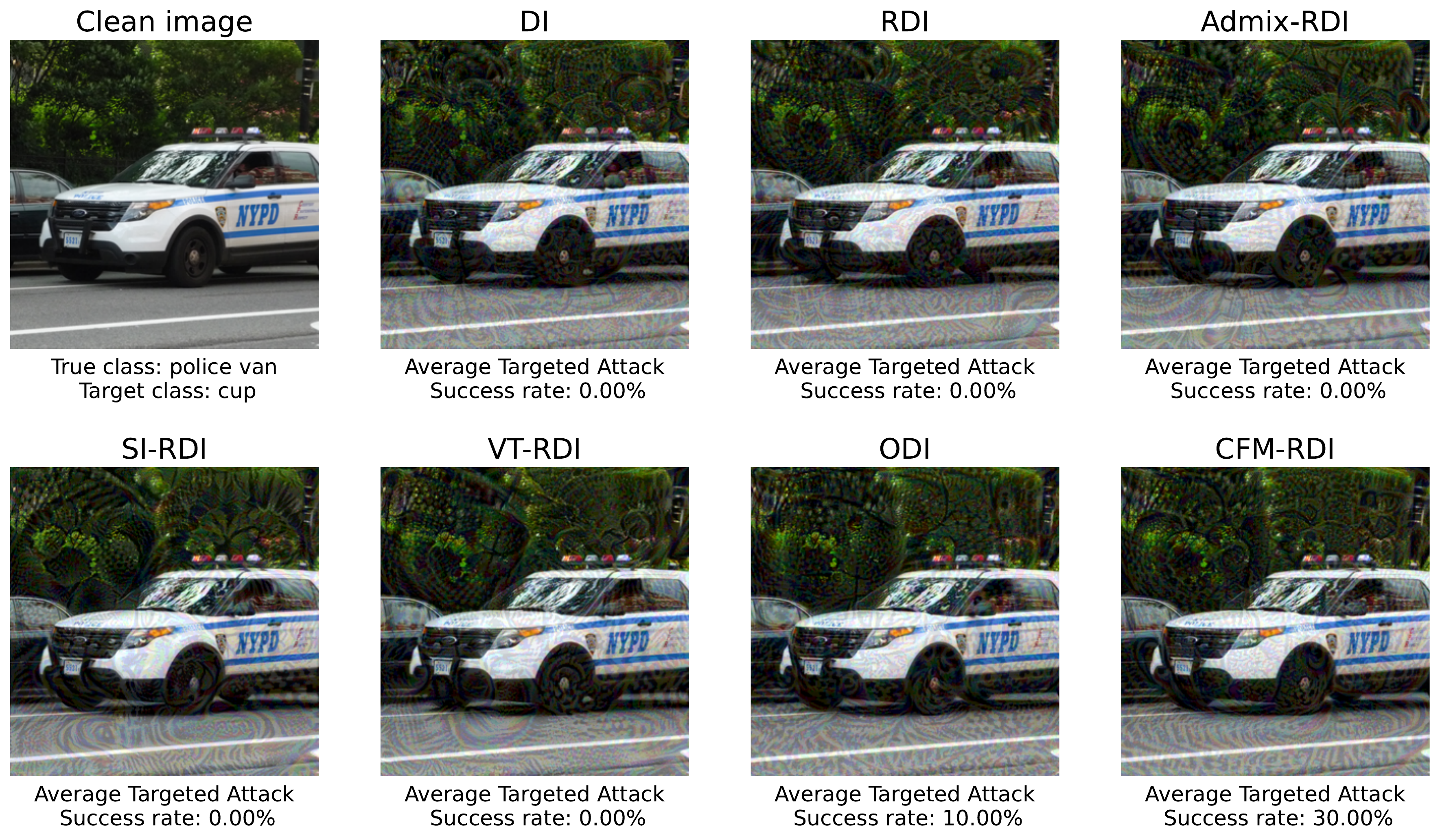}
         \caption{Visualization of generated adversarial examples. The source model is Inc-v3. 
Each average targeted attack success rate was calculated over the ten carefully selected target models, which are more difficult to confuse. For example, an average targeted attack success rate of 50\% means that 5 out of 10 target models recognize the adversarial example as the target class.}
        \label{fig:fig6}
\end{figure*}

Figure \ref{fig:fig1}, \ref{fig:fig2}, \ref{fig:fig3}, \ref{fig:fig4}, \ref{fig:fig5} and \ref{fig:fig6} visualize the generated adversarial examples for qualitative comparison. We denoted the true and target classes below the clean images and computed the average attack success rates over the ten carefully selected pre-trained target models listed in Table \ref{tab:table4}. Note that all adversarial perturbations are constrained by the $\ell_\infty$-norm (\ie $||\bm{\delta}||_\infty\leq\epsilon$ where we used $\epsilon=16/255$). 
\subsection{Extended Experimental Results With Additional Source Models and Baselines}
Table \ref{tab:table1} and Table \ref{tab:table2} show the extended experimental results on the ImageNet-Compatible dataset with additional source models, \ie adv-RN-50 and DN-121 in Table \ref{tab:table1} and RN-50 and DN-121 in Table \ref{tab:table2}. 
For the additional source models, we used the same hyperparameters of CFM as in RN-50 (\ie $\alpha_{max}=0.75$ and $p=0.1$). We also included the results of Admix with the number of scale copies of 5 (\ie $m_1=5$ in \cite{wang2021admix}) and SI-CFM-RDI for more comprehensive comparisons. The Admix$_{m_1=5}$ follows the original setting of the Admix \cite{wang2021admix}, which utilizes the SI technique in its internal loops.

\subsection{Extended Experimental Results on the CIFAR-10 dataset}
Table \ref{tab:table3} shows the extended experimental results on the CIFAR-10 dataset, which additionally include the results of Admix$_{m_1=5}$ and SI-CFM-RDI with different source models (Inc-v3, VGG-16, and DN-121) for more comprehensive comparisons. 

\subsection{Experimental Results of Combined Attacks With Multiple Techniques}
Table \ref{tab:table4} shows the experimental results of various combinations of multiple attack techniques. The results demonstrate that CFM is compatible with existing attack methods, and various combinations with CFM can further improve the transferability of adversarial examples.

\subsection{Experimental Results with Different Mixing Hyperparameters}
Table \ref{tab:table4} shows the experimental results on how the transfer success rates vary by changing the values of the mixing probability $p$ and the upper bound of mixing ratios $\alpha_{max}$. In this experiment, we used adv-RN-50 as the source model and evaluated the transfer success rates on the carefully selected ten target models. CFM achieves the highest success rate when $p=0.1$ and $\alpha_{max}=0.75$, but it also achieves comparable attack success rates at other values. This indicates that CFM is not very sensitive to the changes in hyperparameters and can achieve consistent performance improvement. 


\begin{table*}[ht] 
\resizebox{0.97\textwidth}{!}{%
\begin{tabular}{lccccccccccc}
\toprule[0.15em]
\textbf{\begin{tabular}[c]{@{}l@{}}Source : \\ RN-50\end{tabular}} & \multicolumn{10}{c}{Target model} &  \\\cmidrule(l){2-11}
Attack & VGG-16 & RN-18 & RN-50 & DN-121 & Xcep & MB-v2 & EF-B0 & IR-v2 & Inc-v3 & Inc-v4 & Avg. \\ \midrule
DI & 62.5 & 56.6 & \BL98.9 & 72.3 & 5.7 & 28.2 & 29.3 & 4.5 & 9.2 & 9.9 & 37.7 \\
RDI & 65.4 & 71.8 & 98.0 & 81.3 & 13.1 & 46.6 & 46.6 & 16.8 & 30.7 & 23.9 & 49.4 \\
Admix$_{m_1=1}$-RDI & 74.2 & 80.7 & 98.7 & 86.8 & 20.9 & 59.4 & 56.1 & 26.7 & 42.7 & 34.1 & 58.0 \\
Admix$_{m_1=5}$-RDI & 75.2 & 83.0 & 98.4 & 89.6 & 36.5 & 64.7 & 66.4 & 44.7 & 62.5 & 50.5 & 67.2 \\
SI-RDI & 70.5 & 79.8 & 98.8 & 88.9 & 29.5 & 56.2 & 66.2 & 37.9 & 56.4 & 43.6 & 62.8 \\
VT-RDI & 68.8 & 78.7 & 98.2 & 82.5 & 27.9 & 54.5 & 56.1 & 32.8 & 45.8 & 37.9 & 58.3 \\
ODI & 78.3 & 77.1 & 97.6 & 87.0 & 43.8 & 67.3 & 70.0 & 49.5 & 65.9 & 55.4 & 69.2 \\
\rowcolor{Gray}CFM-RDI & 84.7 & 88.4 & 98.4 & 90.3 & 51.1 & 81.5 & 78.8 & 48.0 & 65.5 & 59.3 & 74.6 \\
\rowcolor{Gray}SI-CFM-RDI & \BL85.9 & \BL88.5 & 98.4 & \BL92.3 & \BL62.5 & \BL81.6 & \BL82.7 & \BL61.5 & \BL74.5 & \BL69.7 & \BL79.8
\\ \midrule
\textbf{\begin{tabular}[c]{@{}l@{}}Source : \\ adv-RN-50\end{tabular}} & \multicolumn{10}{c}{Target model} &  \\\cmidrule(l){2-11}
Attack & VGG-16 & RN-18 & RN-50 & DN-121 & Xcep & MB-v2 & EF-B0 & IR-v2 & Inc-v3 & Inc-v4 & Avg. \\ \midrule
DI & 65.3 & 81.5 & \BL91.5 & 87.0 & 32.6 & 62.5 & 68.8 & 36.9 & 55.3 & 42.2 & 62.4 \\
RDI & 59.7 & 83.5 & 90.7 & 85.9 & 39.7 & 67.0 & 68.8 & 44.2 & 62.4 & 45.1 & 64.7 \\
Admix$_{m_1=1}$-RDI  & 62.7 & 83.0 & 90.3 & 86.6 & 46.9 & 71.8 & 72.4 & 48.8 & 66.3 & 53.0 & 68.2 \\
Admix$_{m_1=5}$-RDI  & 54.4 & 81.0 & 86.0 & 81.8 & 48.8 & 68.0 & 68.5 & 50.7 & 68.3 & 52.9 & 66.0 \\
SI-RDI & 53.9 & 79.4 & 87.1 & 83.8 & 46.6 & 66.5 & 69.5 & 52.0 & 69.1 & 52.2 & 66.0 \\
VT-RDI & 54.0 & 76.8 & 84.7 & 81.2 & 38.5 & 60.3 & 58.7 & 42.7 & 56.1 & 44.9 & 59.8 \\
ODI & 62.0 & 77.6 & 84.3 & 85.0 & 56.3 & 66.9 & 73.0 & 61.1 & 71.9 & 60.0 & 69.8 \\
\rowcolor{Gray}CFM-RDI & \BL76.7 & \BL86.3 & 90.9 & \BL87.6 & \BL67.1 & \BL82.4 & \BL83.4 & \BL64.7 & \BL77.1 & \BL67.4 & \BL78.4 \\
\rowcolor{Gray}SI-CFM-RDI & 70.0 & 82.3 & 86.8 & 85.7 & 63.4 & 79.2 & 79.4 & 61.8 & 76.2 & 63.9 & 74.9 \\\midrule
\textbf{\begin{tabular}[c]{@{}l@{}}Source : \\ Inc-v3\end{tabular}} & \multicolumn{10}{c}{Target model} &  \\\cmidrule(l){2-11}
Attack & VGG-16 & RN-18 & RN-50 & DN-121 & Xcep & MB-v2 & EF-B0 & IR-v2 & Inc-v3 & Inc-v4 & Avg. \\ \midrule
DI & 2.9 & 2.4 & 3.4 & 5.0 & 1.9 & 1.8 & 3.7 & 3.0 & \BL99.2 & 4.2 & 12.8 \\
RDI & 3.5 & 3.8 & 4.0 & 7.0 & 3.1 & 3.0 & 5.9 & 6.3 & 98.7 & 7.1 & 14.2 \\
Admix$_{m_1=1}$-RDI & 6.3 & 6.5 & 8.8 & 12.8 & 6.0 & 6.1 & 10.9 & 12.2 & 98.7 & 13.6 & 18.2 \\
Admix$_{m_1=5}$-RDI & 4.4 & 9.0 & 8.3 & 13.3 & 8.2 & 6.5 & 12.0 & 14.8 & 98.5 & 16.3 & 19.1 \\
SI-RDI & 4.0 & 5.2 & 5.7 & 11.0 & 6.3 & 4.6 & 8.2 & 11.6 & 98.8 & 12.1 & 16.8 \\
VT-RDI & 5.9 & 8.9 & 9.4 & 13.2 & 7.4 & 5.9 & 9.8 & 12.3 & 98.7 & 14.7 & 18.6 \\
ODI & 14.3 & 14.9 & 16.7 & 32.3 & 20.3 & 13.7 & 25.3 & 26.4 & 95.6 & 31.6 & 29.1 \\
\rowcolor{Gray}CFM-RDI & 22.9 & 26.8 & 26.2 & 39.1 & 34.1 & 27.1 & 38.6 & 36.2 & 95.9 & 44.8 & 39.2 \\
\rowcolor{Gray}SI-CFM-RDI & \BL24.4 & \BL36.3 & \BL32.3 & \BL51.1 & \BL44.8 & \BL30.9 & \BL45.7 & \BL52.0 & 97.5 & \BL55.4 & \BL47.0 \\
\midrule
\textbf{\begin{tabular}[c]{@{}l@{}}Source : \\ DN-121\end{tabular}} & \multicolumn{10}{c}{Target model} &  \\\cmidrule(l){2-11}
Attack & VGG-16 & RN-18 & RN-50 & DN-121 & Xcep & MB-v2 & EF-B0 & IR-v2 & Inc-v3 & Inc-v4 & Avg. \\ \midrule
DI & 37.4 & 28.7 & 44.4 & \BL98.7 & 5.2 & 13.1 & 18.7 & 4.3 & 7.1 & 8.3 & 26.6 \\
RDI & 42.1 & 48.8 & 55.7 & 98.5 & 10.1 & 21.0 & 29.0 & 12.8 & 20.8 & 18.8 & 35.8 \\
Admix$_{m_1=1}$-RDI & 53.2 & 60.7 & 67.6 & 98.3 & 17.8 & 31.5 & 39.4 & 20.1 & 31.1 & 26.5 & 44.6 \\
Admix$_{m_1=5}$-RDI & 49.6 & 60.4 & 65.3 & 98.6 & 21.6 & 34.8 & 43.5 & 28.9 & 41.0 & 34.3 & 47.8 \\
SI-RDI & 45.4 & 53.0 & 60.1 & 98.6 & 16.1 & 27.8 & 37.3 & 22.0 & 34.3 & 25.8 & 42.0 \\
VT-RDI & 47.7 & 56.7 & 62.1 & 98.6 & 20.3 & 28.7 & 36.9 & 25.4 & 31.5 & 27.2 & 43.5 \\
ODI & 64.2 & 64.2 & 71.7 & 98.0 & 31.4 & 45.9 & 56.1 & 39.8 & 52.8 & 45.9 & 57.0 \\
\rowcolor{Gray}CFM-RDI & 76.2 & 79.0 & 83.9 & 97.8 & 41.1 & 62.5 & 68.6 & 43.6 & 56.1 & 53.8 & 66.3 \\
\rowcolor{Gray}SI-CFM-RDI & \BL77.2 & \BL81.2 & \BL85.4 & 97.8 & \BL49.7 & \BL67.8 & \BL74.8 & \BL53.8 & \BL67.9 & \BL59.7 & \BL71.5 \\
\bottomrule
\end{tabular}%
}
\caption{Extended experimental results on targeted attack success rates (\%) against the ten target models on the ImageNet-Compatible dataset.}
\label{tab:table1}
\end{table*}

\begin{table*}[ht] 
\centering
\resizebox{0.72\textwidth}{!}{%
\begin{tabular}{lcccccccc}
 \toprule[0.15em]
\textbf{\begin{tabular}[c]{@{}l@{}}Source : \\ RN-50\end{tabular}} & \multicolumn{6}{c}{Target model} & \multicolumn{1}{l}{} & \\\cmidrule(l){2-7}
Attack & \begin{tabular}[c]{@{}c@{}}adv-\\ RN-50\end{tabular} & ViT & LeViT & ConViT & Twins & PiT & Avg. & \begin{tabular}[c]{@{}c@{}}Computation time\\ per image (sec)\end{tabular} \\\midrule
DI & 10.9 & 0.1 & 3.6 & 0.3 & 1.3 & 1.5 & 2.9 & 3.73 \\
RDI & 34.8 & 0.7 & 13.1 & 1.9 & 5.9 & 6.8 & 10.5 & 3.29 \\
Admix$_{m_1=1}$-RDI & 52.4 & 1.3 & 22.5 & 2.5 & 8.5 & 8.4 & 15.9 & 9.73 \\
Admix$_{m_1=5}$-RDI& 68.6 & 4.0 & 36.3 & 7.7 & 18.7 & 20.0 & 25.9 & 49.19 \\
SI-RDI & 59.9 & 2.9 & 29.4 & 6.3 & 15.5 & 17.9 & 22.0 & 16.16 \\
VT-RDI & 64.2 & 2.9 & 28.1 & 5.2 & 15.0 & 14.0 & 21.6 & 19.83 \\
ODI & 64.7 & 5.1 & 37.0 & 10.7 & 20.1 & 29.1 & 27.8 & 9.05 \\
\rowcolor{Gray}CFM-RDI & 75.5 & 4.3 & 46.1 & 8.9 & 25.2 & 24.7 & 30.8 & 3.72 \\
\rowcolor{Gray}SI-CFM-RDI & \BL80.8 & \BL12.4 & \BL60.1 &\BL 16.7 & \BL39.7 & \BL43.3 & \BL42.2 & 18.34 
\\\midrule
\textbf{\begin{tabular}[c]{@{}l@{}}Source : \\ adv-RN-50\end{tabular}} & \multicolumn{6}{c}{Target model} & \multicolumn{1}{l}{} & \\\cmidrule(l){2-7}
Attack & \begin{tabular}[c]{@{}c@{}}adv-\\ RN-50\end{tabular} & ViT & LeViT & ConViT & Twins & PiT & Avg. & \begin{tabular}[c]{@{}c@{}}Computation time\\ per image (sec)\end{tabular} \\\midrule
DI & \BL98.9 & 5.7 & 36.9 & 10.1 & 19.2 & 20.5 & 31.9 & 3.77 \\
RDI & 98.8 & 10.8 & 49.5 & 19.9 & 29.4 & 35.8 & 40.7 & 3.29 \\
Admix$_{m_1=1}$-RDI & \BL98.9 & 12.1 & 55.5 & 23.1 & 32.4 & 38.9 & 43.5 & 9.86 \\
Admix$_{m_1=5}$-RDI & 98.4 & 19.7 & 56.4 & 34.1 & 36.2 & 49.4 & 49.0 & 49.19 \\
SI-RDI & 98.7 & 19.4 & 57.6 & 35.3 & 35.2 & 52.1 & 49.7 & 16.34 \\
VT-RDI & 98.5 & 10.6 & 46.3 & 20.0 & 27.1 & 34.4 & 39.5 & 19.83 \\
ODI & 97.3 & 22.2 & 57.7 & 38.8 & 40.0 & 54.9 & 51.8 & 9.04 \\
\rowcolor{Gray}CFM-RDI & 98.3 & 29.5 & \BL69.8 & 41.8 & \BL52.7 & 59.8 & 58.6 & 3.74 \\
\rowcolor{Gray}SI-CFM-RDI & 98.2 & \BL33.1 & 68.9 & \BL46.6 & 52.2 & \BL61.9 & \BL60.1 & 18.47 \\\midrule
\textbf{\begin{tabular}[c]{@{}l@{}}Source : \\ Inc-v3\end{tabular}} & \multicolumn{6}{c}{Target model} & \multicolumn{1}{l}{} & \\\cmidrule(l){2-7}
Attack & \begin{tabular}[c]{@{}c@{}}adv-\\ RN-50\end{tabular} & ViT & LeViT & ConViT & Twins & PiT & Avg. & \begin{tabular}[c]{@{}c@{}}Computation time\\ per image (sec)\end{tabular} \\\midrule
DI & 0.2 & 0.1 & 0.3 & 0.0 & 0.0 & 0.1 & 0.1 & 2.84 \\
RDI & 0.8 & 0.2 & 1.8 & 0.2 & 0.4 & 0.7 & 0.7 & 2.47 \\
Admix$_{m_1=1}$-RDI & 2.0 & 0.1 & 4.1 & 0.6 & 1.4 & 1.4 & 1.6 & 7.27 \\
Admix$_{m_1=5}$-RDI & 5.0 & 0.8 & 6.4 & 1.6 & 1.6 & 3.9 & 3.2 & 36.30 \\
SI-RDI & 2.0 & 0.3 & 4.1 & 0.9 & 0.7 & 3.2 & 1.9 & 12.23 \\
VT-RDI & 3.2 & 0.4 & 5.2 & 0.8 & 1.6 & 1.8 & 2.2 & 14.74 \\
ODI & 6.5 & 0.8 & 12.4 & 1.7 & 3.5 & 6.7 & 5.3 & 6.74 \\
\rowcolor{Gray}CFM-RDI & 8.6 & 2.1 & 21.9 & 3.2 & 6.1 & 11.6 & 8.9 & 2.96 \\
\rowcolor{Gray}SI-CFM-RDI & \BL19.3 & \B6.1 & \BL33.7 & \B6.8 & \BL12.4 & \BL22.5 & \BL16.8 & 14.63
\\\midrule
\textbf{\begin{tabular}[c]{@{}l@{}}Source : \\ DN-121\end{tabular}} & \multicolumn{6}{c}{Target model} & \multicolumn{1}{l}{} & \\\cmidrule(l){2-7}
Attack & \begin{tabular}[c]{@{}c@{}}adv-\\ RN-50\end{tabular} & ViT & LeViT & ConViT & Twins & PiT & Avg. & \begin{tabular}[c]{@{}c@{}}Computation time\\ per image (sec)\end{tabular} \\\midrule
DI & 3.2 & 0.2 & 3.0 & 0.4 & 1.0 & 1.1 & 1.5 & 3.62 \\
RDI & 10.1 & 0.8 & 8.5 & 1.3 & 3.7 & 4.5 & 4.8 & 3.22 \\
Admix-RDI & 19.2 & 1.0 & 14.7 & 1.7 & 6.8 & 7.4 & 8.5 & 9.46 \\
SI-Admix-RDI & 26.7 & 2.4 & 21.8 & 3.4 & 10.5 & 14.2 & 13.2 & 46.61 \\
SI-RDI & 19.2 & 2.0 & 16.1 & 2.4 & 8.2 & 11.7 & 9.9 & 15.65 \\
VT-RDI & 26.6 & 2.2 & 19.2 & 3.5 & 8.3 & 11.7 & 11.9 & 18.87 \\
ODI & 35.6 & 3.3 & 26.9 & 7.4 & 14.7 & 21.9 & 18.3 & 9.06 \\
\rowcolor{Gray}CFM-RDI & 43.2 & 3.6 & 32.8 & 6.4 & 17.3 & 21.1 & 20.7 & 3.69 \\
\rowcolor{Gray}SI-CFM-RDI & \BL 54.3 & \B 8.0 & \BL46.5 & \BL11.8 & \BL28.4 & \BL35.5 & \BL30.8 & 18.18 \\
\bottomrule
\end{tabular}%
}

\caption{Extended experimental results of targeted attack success rates (\%) against one adversarially trained model and five Transformer-based classifiers with the ImageNet-Compatible dataset. We also report the average computation time to construct an adversarial example.}
\label{tab:table2}
\end{table*}

\begin{table*}[ht]
\centering
\resizebox{\textwidth}{!}{%

\begin{tabular}{lccccccccccc}
\toprule[0.15em]
\textbf{\begin{tabular}[c]{@{}l@{}}Source : \\ RN-50\end{tabular}} & \multicolumn{9}{c}{Target model} & \multicolumn{1}{l}{} &  \\\cmidrule(l){2-10}
\multirow{2}{*}{Attack} & \multirow{2}{*}{VGG-16} & \multirow{2}{*}{RN-18} & \multirow{2}{*}{MB-v2} & \multirow{2}{*}{Inc-v3} & \multirow{2}{*}{DN-121} & \multicolumn{4}{c}{ens3-RN-20} & \multirow{2}{*}{Avg.} & \multirow{2}{*}{\begin{tabular}[c]{@{}c@{}}Computation time\\ per image (sec)\end{tabular}} \\
 &  &  &  &  &  & Baseline & ADP & GAL & DVERGE &  &  \\ \midrule
DI & 66.4 & 71.5 & 62.7 & 71.1 & 84.2 & 77.9 & 56.5 & 14.3 & 15.6 & 57.8 & 0.64 \\
RDI & 66.4 & 70.9 & 64.1 & 73.4 & 82.8 & 76.3 & 55.8 & 13.5 & 14.9 & 57.6 & 0.59 \\
SI-RDI & 72.9 & 76.3 & 77.1 & 77.0 & 84.7 & 81.2 & 65.5 & 20.0 & 22.4 & 64.1 & 3.17 \\
VT-RDI & 89.8 & 87.1 & 92.6 & 92.9 & 93.7 & 94.4 & 82.3 & 24.3 & 31.3 & 76.5 & 3.82 \\
Admix$_{m_1=1}$-RDI & 74.2 & 78.8 & 76.2 & 82.7 & 89.2 & 85.2 & 66.4 & 17.3 & 18.4 & 65.4 & 1.98 \\
Admix$_{m_1=5}$-RDI & 79.9 & 82.3 & 81.3 & 83.4 & 90.0 & 86.0 & 69.7 & 22.8 & 25.6 & 69.0 & 9.06 \\
\rowcolor{Gray}CFM-RDI & 98.3 & 97.7 & 99.0 & \BL99.0 & \BL99.2 & \BL98.8 & 97.2 & 54.9 & 59.3 & 89.3 & 0.72 \\
\rowcolor{Gray}SI-CFM-RDI & \BL98.5 & \BL98.1 & \BL99.2 & 98.9 & \BL99.2 & \BL98.8 & \BL97.3 & \BL61.3 & \BL65.8 & \BL90.8 & 5.02
\\        \midrule
\textbf{\begin{tabular}[c]{@{}l@{}}Source : \\ Inc-v3\end{tabular}} & \multicolumn{9}{c}{Target model} & \multicolumn{1}{l}{} &  \\\cmidrule(l){2-10}
\multirow{2}{*}{Attack} & \multirow{2}{*}{VGG-16} & \multirow{2}{*}{RN-18} & \multirow{2}{*}{MB-v2} & \multirow{2}{*}{Inc-v3} & \multirow{2}{*}{DN-121} & \multicolumn{4}{c}{ens3-RN-20} & \multirow{2}{*}{Avg.} & \multirow{2}{*}{\begin{tabular}[c]{@{}c@{}}Computation time\\ per image (sec)\end{tabular}} \\
 &  &  &  &  &  & Baseline & ADP & GAL & DVERGE &  &  \\ \midrule
DI & 22.8 & 12.8 & 32.1 & 78.7 & 14.7 & 32.8 & 21.7 & 3.8 & 3.5 & 24.8 & 1.74 \\
RDI & 21.3 & 14.7 & 33.9 & 86.7 & 16.6 & 37.6 & 22.3 & 4.4 & 5.0 & 26.9 & 1.88 \\
SI-RDI & 43.1 & 28.5 & 51.4 & \BL99.8 & 30.0 & 60.9 & 46.5 & 11.9 & 9.1 & 42.4 & 8.54 \\
VT-RDI & 53.7 & 31.1 & 72.0 & 93.3 & 38.5 & 69.6 & 55.7 & 8.9 & 9.4 & 48.0 & 10.43 \\
Admix$_{m_1=1}$-RDI & 29.6 & 16.9 & 43.3 & 90.5 & 19.9 & 47.4 & 30.9 & 5.3 & 5.1 & 32.1 & 5.60 \\
Admix$_{m_1=5}$-RDI & 47.2 & 32.0 & 58.1 & 99.7 & 34.8 & 64.0 & 50.8 & 14.9 & 11.2 & 45.9 & 30.94 \\
\rowcolor{Gray}CFM-RDI & 45.3 & 29.1 & 57.2 & 94.7 & 33.9 & 55.7 & 42.2 & 8.8 & 8.3 & 41.7 & 2.02 \\
\rowcolor{Gray}SI-CFM-RDI & \BL62.3 & \BL45.5 & \BL67.5 & 99.5 & \BL49.9 & \BL71.6 & \BL60.9 & \BL19.1 & \BL15.5 & \BL54.6 & 8.84 \\

        \midrule
\textbf{\begin{tabular}[c]{@{}l@{}}Source : \\ DN-121\end{tabular}} & \multicolumn{9}{c}{Target model} & \multicolumn{1}{l}{} &  \\\cmidrule(l){2-10}
\multirow{2}{*}{Attack} & \multirow{2}{*}{VGG-16} & \multirow{2}{*}{RN-18} & \multirow{2}{*}{MB-v2} & \multirow{2}{*}{Inc-v3} & \multirow{2}{*}{DN-121} & \multicolumn{4}{c}{ens3-RN-20} & \multirow{2}{*}{Avg.} & \multirow{2}{*}{\begin{tabular}[c]{@{}c@{}}Computation time\\ per image (sec)\end{tabular}} \\
 &  &  &  &  &  & Baseline & ADP & GAL & DVERGE &  &  \\ \midrule
DI & 44.3 & 46.5 & 39.0 & 45.6 & 92.8 & 41.0 & 30.9 & 9.3 & 9.1 & 39.8 & 0.95 \\
RDI & 43.3 & 44.8 & 37.6 & 46.0 & 92.9 & 38.7 & 28.1 & 9.4 & 9.9 & 39.0 & 1.01 \\
SI-RDI & 51.3 & 47.4 & 48.1 & 52.6 & 98.3 & 46.4 & 37.5 & 11.0 & 11.1 & 44.9 & 6.33 \\
VT-RDI & 67.9 & 62.9 & 67.1 & 69.2 & 91.3 & 61.3 & 52.6 & 13.9 & 16.8 & 55.9 & 7.45 \\
Admix$_{m_1=1}$-RDI & 50.7 & 53.9 & 45.7 & 52.8 & 93.1 & 48.4 & 37.0 & 9.9 & 10.4 & 44.7 & 2.86 \\
Admix$_{m_1=5}$-RDI & 62.6 & 58.5 & 57.9 & 62.8 & 98.3 & 56.4 & 44.1 & 14.0 & 13.8 & 52.0 & 14.70 \\
\rowcolor{Gray}CFM-RDI & 97.0 & \BL96.5 & 95.9 & 97.6 & \textbf{100.0} & 95.8 & 91.9 & 43.4 & 45.1 & 84.8 & 1.28 \\
\rowcolor{Gray}SI-CFM-RDI & \BL97.3 & 96.2 & \BL97.1 & \BL97.7 & 99.6 & \BL96.2 & \BL92.6 & \BL49.1 & \BL52.0 & \BL86.4 & 7.44\\

\bottomrule
\end{tabular}%
}
\caption{Targeted attack success rates (\%) against nine target models, including four ensemble-based defensive models on the CIFAR-10 dataset. We also evaluated the average computation time for crafting an adversarial example.}
\label{tab:table3}
\vspace{-0.4cm}
\end{table*}

\begin{table*}[ht]
\centering
\resizebox{\textwidth}{!}{
\begin{tabular}{lcccccccccccc}
\toprule[0.15em]
\textbf{\begin{tabular}[c]{@{}l@{}}Source : RN-50\end{tabular}} & \multicolumn{10}{c}{Target model} & \multicolumn{1}{l}{} & \multicolumn{1}{l}{} \\\cmidrule(l){2-11}
Attack & Xcep & MB-v2 & EF-B0 & IR-v2 & Inc-v4 & ViT & LeViT & ConViT & Twins & PiT & Avg. & \multicolumn{1}{l}{\begin{tabular}[c]{@{}c@{}}Comput. time\\ per image (sec)\end{tabular}} \\ \midrule
None (-MI-TI) & 0.6 & 2.9 & 1.6 & 0.1 & 0.5 & 0.0 & 0.0 & 0.0 & 0.0 & 0.0 & 0.6 & 3.27 \\
RDI & 13.1 & 46.6 & 46.6 & 16.8 & 23.9 & 0.7 & 13.1 & 1.9 & 5.9 & 6.8 & 17.5 & 3.29 \\
SI-RDI & 29.5 & 56.2 & 66.2 & 37.9 & 43.6 & 2.9 & 29.4 & 6.3 & 15.5 & 17.9 & 30.5 & 16.16 \\
VT-RDI & 27.9 & 54.5 & 56.1 & 32.8 & 37.9 & 2.9 & 28.1 & 5.2 & 15.0 & 14.0 & 27.4 & 19.83 \\
ODI & 43.8 & 67.3 & 70.0 & 49.5 & 55.4 & 5.1 & 37.0 & 10.7 & 20.1 & 29.1 & 38.8 & 9.05 \\
ODI-RDI & 45.8 & 65.8 & 69.0 & 48.2 & 51.4 & 6.2 & 41.9 & 11.8 & 22.8 & 31.9 & 39.5 & 9.77 \\
Admix$_{m_1=1}$-RDI& 20.9 & 59.4 & 56.1 & 26.7 & 34.1 & 1.3 & 22.5 & 2.5 & 8.5 & 8.4 & 24.0 & 9.73 \\
Admix$_{m_1=5}$-RDI & 36.5 & 64.7 & 66.4 & 44.7 & 50.5 & 4.0 & 36.3 & 7.7 & 18.7 & 20.0 & 35.0 & 49.19 \\
VT-Admix$_{m_1=1}$-RDI & 33.5 & 61.2 & 58.9 & 37.5 & 43.0 & 4.9 & 35.0 & 6.1 & 16.4 & 17.9 & 31.4 & 58.08 \\
\rowcolor{Gray}CFM & 6.3 & 35.2 & 31.9 & 4.9 & 9.4 & 0.0 & 3.1 & 0.2 & 0.8 & 1.2 & 9.3 & 3.35 \\
\rowcolor{Gray}CFM-RDI & 51.1 & 81.5 & 78.8 & 48.0 & 59.3 & 4.3 & 46.1 & 8.9 & 25.2 & 24.7 & 42.8 & 3.72 \\
\rowcolor{Gray}CFM-ODI & 55.1 & 72.5 & 73.4 & 55.4 & 60.7 & 8.6 & 48.7 & \BL16.7 & 30.1 & 39.0 & 46.0 & 9.13 \\
\rowcolor{Gray}SI-CFM-RDI & 62.5 & 81.6 & \BL82.7 & 61.5 & 69.7 & 12.4 & 60.1 & \BL16.7 & 39.7 & 43.3 & 53.0 & 18.34 \\
\rowcolor{Gray}VT-CFM-RDI & 57.3 & 77.4 & 74.6 & 55.2 & 62.0 & 11.2 & 53.0 & 15.7 & 33.6 & 36.3 & 47.6 & 20.69 \\
\rowcolor{Gray}Admix$_{m_1=1}$-CFM-RDI & 56.6 & \BL84.0 & 81.7 & 51.1 & 64.8 & 6.3 & 52.3 & 10.7 & 28.1 & 29.5 & 46.5 & 9.99 \\
\rowcolor{Gray}Admix$_{m_1=5}$-CFM-RDI & \BL65.9 & 81.6 & 82.6 &\BL61.8 & \BL69.9 & \BL12.5 & \BL60.4 & \BL16.7 & \BL40.0 & \BL42.4 & \BL53.4 & 52.57 \\
\midrule
\textbf{\begin{tabular}[c]{@{}l@{}}Source : adv-RN-50\end{tabular}} & \multicolumn{10}{c}{Target model} & \multicolumn{1}{l}{} & \multicolumn{1}{l}{} \\\cmidrule(l){2-11}
Attack & Xcep & MB-v2 & EF-B0 &  IR-v2 & Inc-v4  & ViT & LeViT & ConViT & Twins & PiT & Avg. & \multicolumn{1}{l}{\begin{tabular}[c]{@{}c@{}}Comput. time\\ per image (sec)\end{tabular}} \\ \midrule
None (-MI-TI) & 7.7 & 18.6 & 23.8 & 8.2 & 6.8 & 0.6 & 7.8 & 1.4 & 3.6 & 3.9 & 8.2 & 3.27 \\
RDI & 39.7 & 67.0 & 68.8 & 44.2 & 45.1 & 10.8 & 49.5 & 19.9 & 29.4 & 35.8 & 41.0 & 3.29 \\
SI-RDI & 46.6 & 66.5 & 69.5 & 52.0 & 52.2 & 19.4 & 57.6 & 35.3 & 35.2 & 52.1 & 48.6 & 16.34 \\
VT-RDI & 38.5 & 60.3 & 58.7 & 42.7 & 44.9 & 10.6 & 46.3 & 20.0 & 27.1 & 34.4 & 38.4 & 19.83 \\
ODI & 56.3 & 66.9 & 73.0 & 61.1 & 60.0 & 22.2 & 57.7 & 38.8 & 40.0 & 54.9 & 53.1 & 9.04 \\
ODI-RDI & 52.8 & 65.6 & 68.8 & 57.1 & 56.8 & 25.1 & 57.3 & 39.5 & 38.0 & 53.5 & 51.5 & 9.96 \\
Admix$_{m_1=1}$-RDI & 46.9 & 71.8 & 72.4 & 48.8 & 53.0 & 12.1 & 55.5 & 23.1 & 32.4 & 38.9 & 45.5 & 9.86 \\
Admix$_{m_1=5}$-RDI  & 48.8 & 68.0 & 68.5 & 50.7 & 52.9 & 19.7 & 56.4 & 34.1 & 36.2 & 49.4 & 48.5 & 49.19 \\
VT-Admix$_{m_1=1}$-RDI & 42.8 & 63.4 & 59.2 & 45.1 & 45.4 & 13.6 & 46.7 & 21.9 & 28.2 & 38.1 & 40.4 & 58.08 \\
\rowcolor{Gray}CFM & 54.3 & 80.3 & 80.5 & 50.5 & 57.9 & 11.3 & 51.7 & 18.5 & 32.4 & 33.6 & 47.1 & 3.39 \\
\rowcolor{Gray}CFM-RDI & \BL67.1 & \BL82.4 & \BL83.4 & \BL64.7 & \BL67.4 & 29.5 & \BL69.8 & 41.8 & \BL52.7 & 59.8 & \BL61.9 & 3.74 \\
\rowcolor{Gray}CFM-ODI & 55.4 & 68.6 & 69.9 & 56.2 & 57.3 & 27.6 & 57.7 & 41.7 & 39.7 & 55.8 & 53.0 & 9.12 \\
\rowcolor{Gray}SI-CFM-RDI & 63.4 & 79.2 & 79.4 & 61.8 & 63.9 & \BL33.1 & 68.9 & \BL46.6 & 52.2 & \BL61.9 & 61.0 & 18.47 \\
\rowcolor{Gray}VT-CFM-RDI & 58.4 & 75.7 & 73.2 & 58.1 & 59.4 & 25.4 & 61.3 & 40.0 & 45.2 & 53.7 & 55.0 & 20.71 \\
\rowcolor{Gray}Admix$_{m_1=1}$-CFM-RDI& 63.6 & 81.8 & 81.6 & 62.1 & 64.2 & 29.0 & 67.1 & 39.9 & 49.4 & 57.7 & 59.6 & 9.97 \\
\rowcolor{Gray}Admix$_{m_1=5}$-CFM-RDI & 59.0 & 75.9 & 75.5 & 58.6 & 58.0 & 30.3 & 64.9 & 43.9 & 45.1 & 57.6 & 56.9 & 53.05 \\
             \bottomrule
        \end{tabular}%
}
\caption{Targeted attack success rates (\%) of the combined attacks with multiple techniques against the ten selected target models, which are more difficult to be disturbed. The experiment was conducted on the ImageNet-Compatible dataset.}
\label{tab:table4}
\end{table*}

\begin{table*}[!t]
\centering
\resizebox{0.85\textwidth}{!}{
\begin{tabular}{ccccccccccccc}
\toprule[0.15em]
\multicolumn{2}{c}{Ablation} & \multicolumn{10}{c}{Target model} & \multicolumn{1}{l}{} \\\cmidrule(l){1-2} \cmidrule(l){3-12}
$p$& $\alpha_{max}$ &Xcep & MB-v2 & EF-B0 & IR-v2 & Inc-v4 & ViT & LeViT & ConViT & Twins & PiT & Avg. \\ \midrule
0.05 & 0.5 & 55.2 & 78.7 & 79.2 & 54.4 & 60.9 & 15.7 & 62.8 & 29.1 & 41.0 & 47.9 & 52.5 \\
0.05 & 0.75 & 59.8 & 82.1 & 82.3 & 61.3 & 65.3 & 20.5 & 67.5 & 33.8 & 46.2 & 53.2 & 57.2 \\
0.05 & 1.0 & 63.9 & \BL83.3 & \BL83.9 & 63.0 & \BL68.0 & 24.8 & \BL69.8 & 40.1 & 50.6 & 56.6 & 60.4 \\\midrule
0.1 & 0.5 &61.3 & 81.7 & 80.8 & 62.5 & 64.1 & 22.8 & 69.0 & 37.4 & 46.2 & 54.1 & 58.0 \\
\rowcolor{Gray}0.1 & 0.75 & \BL67.1 & 82.4 & 83.4 & \BL64.7 & 67.4 & \BL29.5 & \BL69.8 & \BL41.8 & \BL52.7 & \BL59.8 & \BL61.9 \\
0.1 & 1.0 &  64.9 & 81.5 & 81.0 & 61.6 & 66.7 & 28.1 & 66.6 & 41.5 & 49.6 & \BL59.8 & 60.1 \\\midrule
0.15 & 0.5 & 64.2 & 82.6 & 81.9 & 63.6 & 67.7 & 27.0 & 68.7 & 41.0 & 50.8 & 58.0 & 60.5  \\
0.15 & 0.75 &  62.6 & 80.4 & 80.2 & 61.1 & 64.0 & 28.7 & 65.2 & 40.9 & 49.5 & 56.7 & 58.9  \\
0.15 & 1.0 & 53.2 & 73.6 & 72.5 & 49.7 & 52.2 & 22.0 & 57.2 & 34.9 & 39.0 & 48.6 & 50.3 \\
\bottomrule
\end{tabular}%
}
\caption{Targeted attack success rates (\%) of CFM-RDI with different mixing probability $p$ and upper bound of mixing ratios $\alpha_{max}$. The source model is adv-RN-50.}
\label{tab:table5}
\end{table*}

{\small
\bibliographystyle{ieee_fullname}
\bibliography{egbib}
}